\documentclass{dissertation}

\usepackage{amsmath,mathtools,empheq}
\usepackage{amsthm}
\usepackage{thmtools}
\usepackage{thm-restate}

\usepackage{longtable}

\usepackage{pdflscape}
\usepackage{adjustbox}

\usepackage{listings}
\usepackage{xcolor}

\definecolor{codegreen}{rgb}{0,0.6,0}
\definecolor{codegray}{rgb}{0.5,0.5,0.5}
\definecolor{codepurple}{rgb}{0.58,0,0.82}
\definecolor{backcolour}{rgb}{0.95,0.95,0.92}

\lstdefinestyle{mystyle}{
  backgroundcolor=\color{backcolour},   commentstyle=\color{codegreen},
  keywordstyle=\color{magenta},
  numberstyle=\tiny\color{codegray},
  stringstyle=\color{codepurple},
  basicstyle=\ttfamily\footnotesize,
  breakatwhitespace=false,         
  breaklines=true,                 
  captionpos=b,                    
  keepspaces=true,                 
  numbers=left,                    
  numbersep=5pt,                  
  showspaces=false,                
  showstringspaces=false,
  showtabs=false,                  
  tabsize=2
}

\usepackage{wrapfig}

\usepackage{enumitem}

\usepackage{tikz}

\usetikzlibrary{calc}
\usepackage{etoolbox}
\robustify
\robustify\addtocounter
\robustify
\robustify

\usetikzlibrary{positioning}
\usetikzlibrary{shapes}

\usetikzlibrary{patterns}
\usetikzlibrary{backgrounds}
\usetikzlibrary{external}
\tikzexternaldisable

\usetikzlibrary{crypto.symbols}
\tikzset{shadows=no}        

\usepackage{forest}
\usepackage{pgfplots}
\usetikzlibrary{crypto.symbols}
\pgfplotsset{compat=newest}
\usetikzlibrary{pgfplots.colormaps}
\usetikzlibrary{pgfplots.groupplots}
\tikzset{shadows=no}

\pgfplotsset{%
    colormap={WhiteRedBlack}{%
        rgb255=(255,255,255)
        rgb255=(255,205,0)
        rgb255=(255,0,0)
        rgb255=(128,0,0)
        rgb255=(0,0,0)
    },
}

\pgfplotsset{%
    colormap={BlueWhiteRed}{%
        rgb255=(0,0,128)
        rgb255=(0,0,255)
        rgb255=(255,255,255)
        rgb255=(255,0,0)
        rgb255=(128,0,0)
    },
}

\definecolor{RoyalBlue}{cmyk}{1,0.5,0,0}
\definecolor{Black}{cmyk}{0,0,0,0}
\definecolor{alertred}{rgb}{0.80,0.12,0.12}
\definecolor{linkgreen}{RGB}{0,166,0}

\def\func/{\mathrm{Func}}
\def\perm/{\mathrm{Perm}}
\def\boolfunc/{\mathrm{BF}}

\makeatletter
\newcommand*{\bottop}{%
  {\mathpalette\@bottop{}}%
}
\newcommand*{\@bottop}[2]{%
  \rlap{$#1\bot\m@th$}
  \top
}
\makeatother

\graphicspath{{./figures/}{./tikzpictures/}}
\pgfplotsset{%
    table/search path={./figures/plots/},
}

\newrobustcmd*{\fullfullcite}{%
    \AtNextCite{%
        \AtEachCitekey{%
            \defcounter{maxnames}{99}%
            \DeclareNameAlias{labelname}{given-family}%
        }%
    }%
    \fullcite
}

\usepackage{pdfpages}

\title{Exposing Hardware Building Blocks to Machine Learning Frameworks}
\author{Yash Akhauri}
\date{\today}

\begin{document}

\includepdf[pages=-, angle=90]{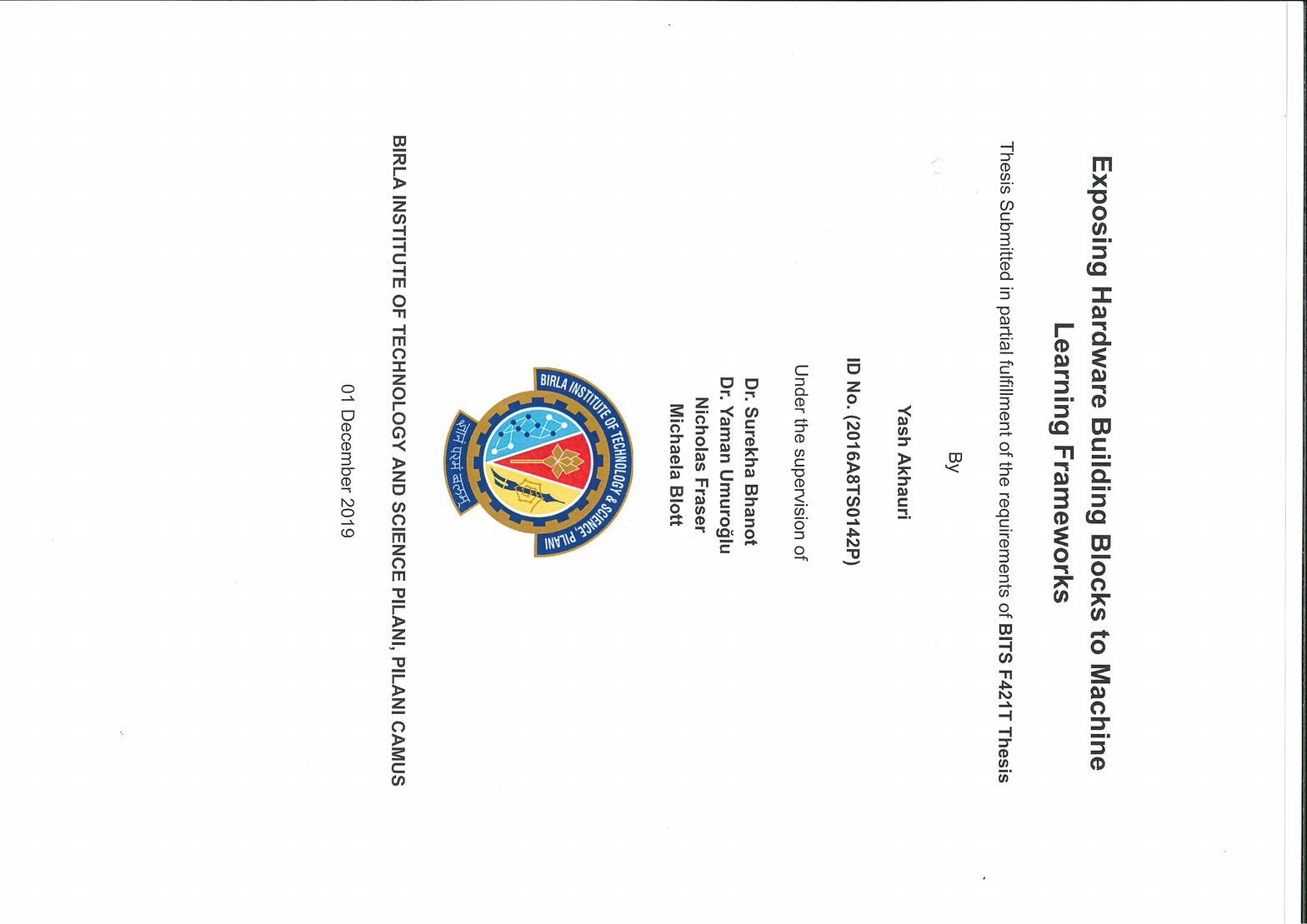}
\includepdf[pages=-, angle=90]{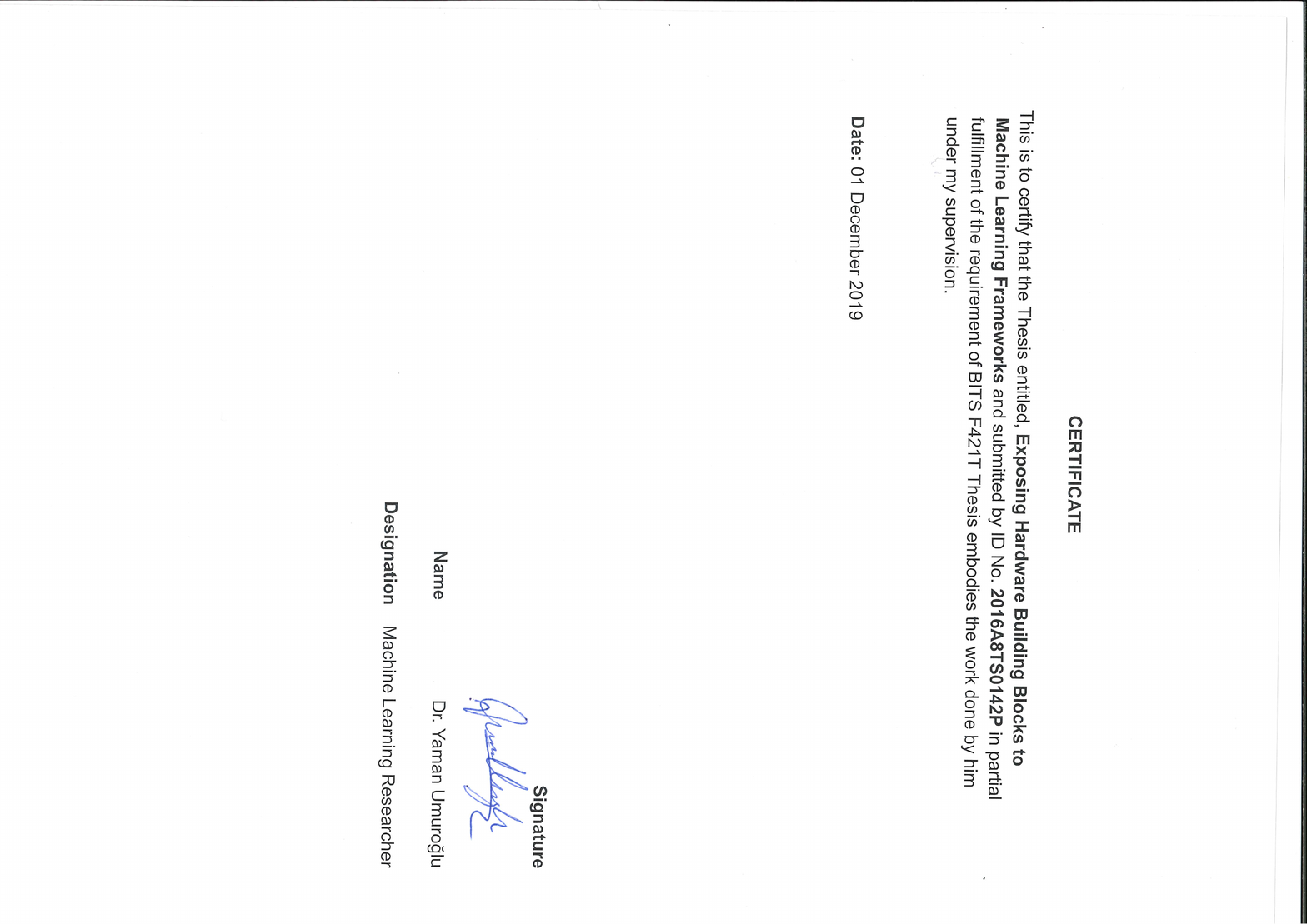}

\frontmatter{}

\cleardoublepage{}

\thispagestyle{empty}
{
    \calccentering{\unitlength}
    \begin{adjustwidth*}{\unitlength}{-\unitlength}
        \raggedleft{}
        {\Huge\color{Burgundy}%
        Exposing Hardware Building Blocks\\
        to Machine Learning Frameworks}\\[\baselineskip]
        {\LARGE%
        Bachelor Thesis}\\[0.1\textheight]
        {\Huge
        Yash Akhauri}\\[\baselineskip]
        {\LARGE%
        01st~December 2019}
        \\[32pt]
        \vfill
        \includegraphics[width=70pt]{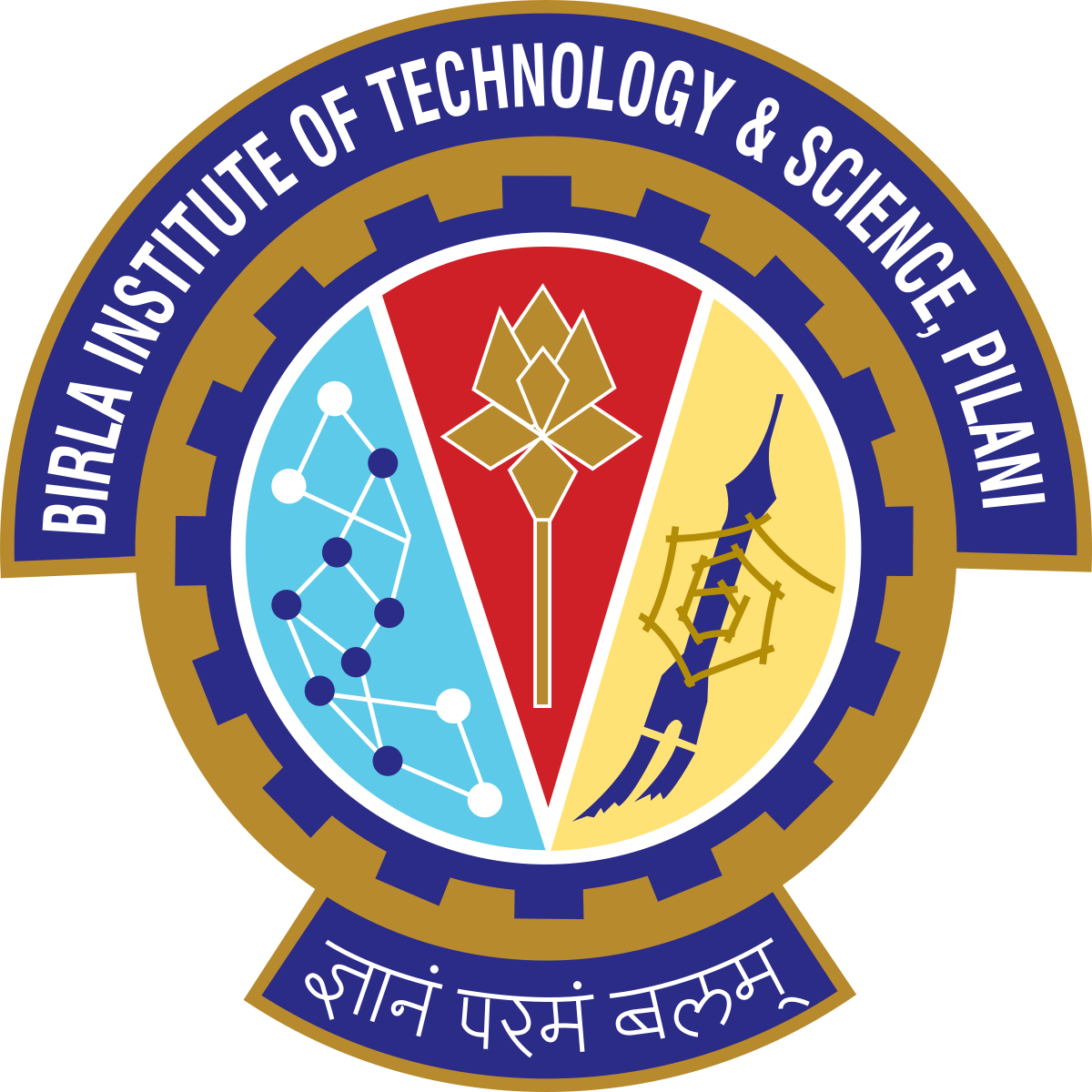} 
        \hfill
        \includegraphics[width=200pt]{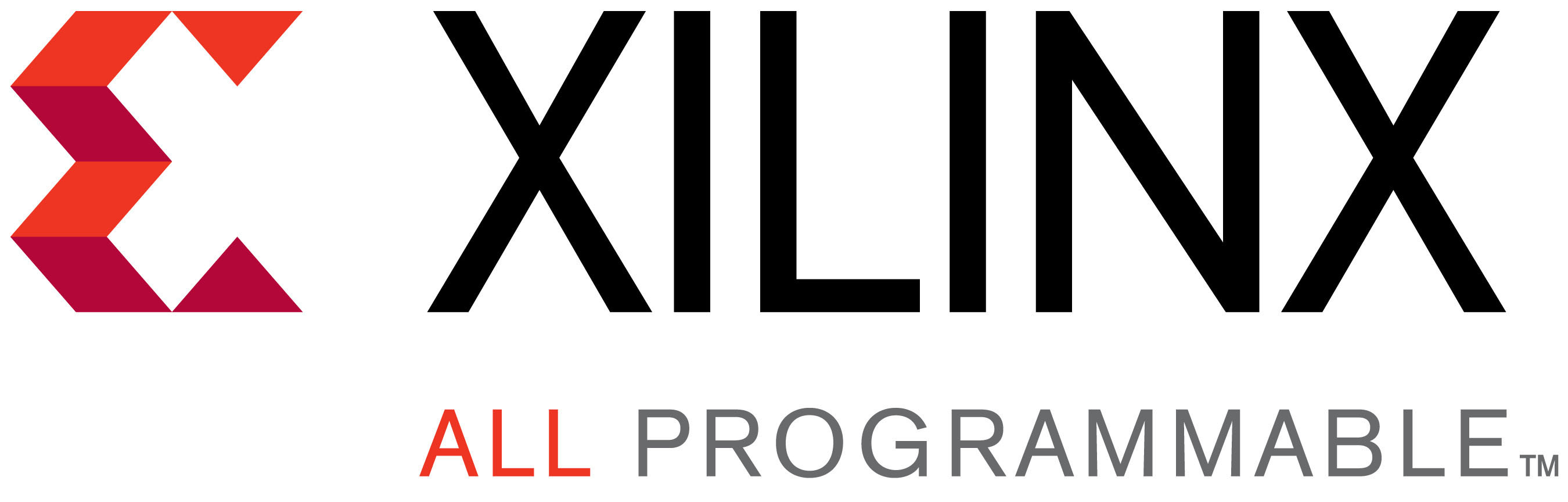}
        \\[32pt]
        \vfill
            {\large%
            Submitted in partial fulfillment of the requirements\\
            for the degree of Bachelor of Engineering\\[\baselineskip]
    
            to the\\[\baselineskip]
    
            Faculty of Electronics and Instrumentation\\
            at Birla Institute of Technology and Science\\[2\baselineskip]
    
            \begin{minipage}{0.5\textwidth}
            \begin{tabular}{lr}
                1st~~Reviewer & Prof.\ Dr. Surekha Bhanot\\
                2nd Reviewer & Dr.\; Yaman Umuroglu\\
                3rd~~Reviewer & Nicholas Fraser\\
                4th~~Reviewer & Michaela Blott
            \end{tabular}
            \end{minipage}
            \hspace*{36pt}
            \vfill
            }
            
        \vspace*{\baselineskip}
    \end{adjustwidth*}
}

\clearpage{}

\thispagestyle{empty}
\hphantom{.}
\vfill

\section*{Imprint}

\textit{Exposing Hardware Building Blocks to Machine Learning Frameworks}\\
Copyright \textcopyright{} 2019 by \theauthor{}.\\
All rights reserved. Printed in India.\\
Published by the Birla Institute of Technology and Science.

\section*{Colophon}

This thesis was typeset using \LaTeX{} and the \texttt{memoir} documentclass.
It is based on Aaron Turon's thesis \emph{Understanding and expressing scalable concurrency}\footnote{\url{https://people.mpi-sws.org/~turon/turon-thesis.pdf}}, itself a mixture of \texttt{classicthesis}\footnote{\url{https://bitbucket.org/amiede/classicthesis/}} by Andr\'e Miede and \texttt{tufte-latex}\footnote{\url{https://github.com/Tufte-LaTeX/tufte-latex}}, based on Edward Tufte's \emph{Beautiful Evidence}.\\[0.5\baselineskip]
The bibliography was processed by Biblatex.
All graphics and plots are made with PGF/Ti\emph{k}Z.\\[0.5\baselineskip]
The body text is set 10/14pt (long primer) on a 26pc measure.
The margin text is set 8/9pt (brevier) on a 12pc measure.
Matthew Carter's \textrm{Charter} acts as both the text and display typeface.
Monospaced text uses Jim Lyles's \texttt{Bitstream Vera Mono} (\enquote{Bera Mono}).

\clearpage{}

\thispagestyle{empty}
\vphantom{.}
\vfill
{%
    \flushright{}
    \emph{If we knew what it was we were doing,\\
          it would not be called research, would it?}\\
    \hfill---Albert Einstein
}
\vfill
\vfill


\chapter*{Abstract}\addcontentsline{toc}{chapter}{Abstract}

There are a plethora of applications that demand high throughput and low latency algorithms leveraging machine learning methods. This need for real time processing can be seen in industries ranging from developing neural network based pre-distortors~\cite{tarver2019design} for enhanced mobile broadband to designing FPGA-based triggers~\cite{duarte2018fast} in major scientific efforts by CERN for particle physics. In this thesis, we explore how niche domains can benefit vastly if we look at neurons as a unique boolean function of the form $f:B^{I} \rightarrow B^{O}$, where $B = \{0,1\}$. We focus on how to design topologies that complement such a view of neurons, how to automate such a strategy of neural network design, and inference of such networks on Xilinx FPGAs.\\

Major hardware borne constraints arise when designing topologies that view neurons as unique boolean functions. Fundamentally, realizing such topologies on hardware asserts a strict limit on the 'fan-in' bits of a neuron due to the doubling of permutations possible with every increment in input bit-length. We address this limit by exploring different methods of implementing sparsity and explore activation quantization. Further, we develop a library that supports training a neural network with custom sparsity and quantization. This library also supports conversion of trained Sparse Quantized networks from PyTorch to VERILOG code which is then synthesized using Vivado, all of which is part of the LogicNet tool-flow. To aid faster prototyping, we also support calculation of the worst-case hardware cost of any given topology. \\

We hope that our insights into the behavior of extremely sparse quantized neural networks are of use to the research community and by extension allow people to use the LogicNet design flow to deploy highly efficient neural networks. We refer the reader to~\cite{umuroglu2020logicnets} and ~\cite{umuroglu2020logicnets_fccm} to read more about this thesis.

\clearpage{}

\tableofcontents*\addcontentsline{toc}{chapter}{Contents}
\listofalgorithms\addcontentsline{toc}{chapter}{List of Algorithms}
\vspace{\baselineskip}

\listoffigures*\addcontentsline{toc}{chapter}{List of Figures}
\vspace{\baselineskip}

\listoftables*\addcontentsline{toc}{chapter}{List of Tables}
\clearpage{}

\cleardoublepage{}


\mainmatter{}
\begin{fullwidth}
\part{Background}
\end{fullwidth}
\cleardoublepage{}
\chapter{Introduction}
\openepigraph{Nanos gigantum humeris insidentes.}{Bernard of Chartres}
\openepigraph{If I have seen further, it is by standing on the shoulders of giants.}{Isaac Newton}

Artificial Intelligence has transformed our industry in the recent past. With every iteration of every conference we hear a buzz around smaller and faster deployments of neural networks, along with a constant push in making the hardware that supports this boom more efficient. From companies investing millions in specialized accelerators (ASICs) for their specific application to hardware companies transforming their business strategy to keep up with the buzz, more (inference) for less (energy and time) seems to be the holy grail of the coming decade. 

\begin{figure}
    \centering
    \includegraphics[width=400pt]{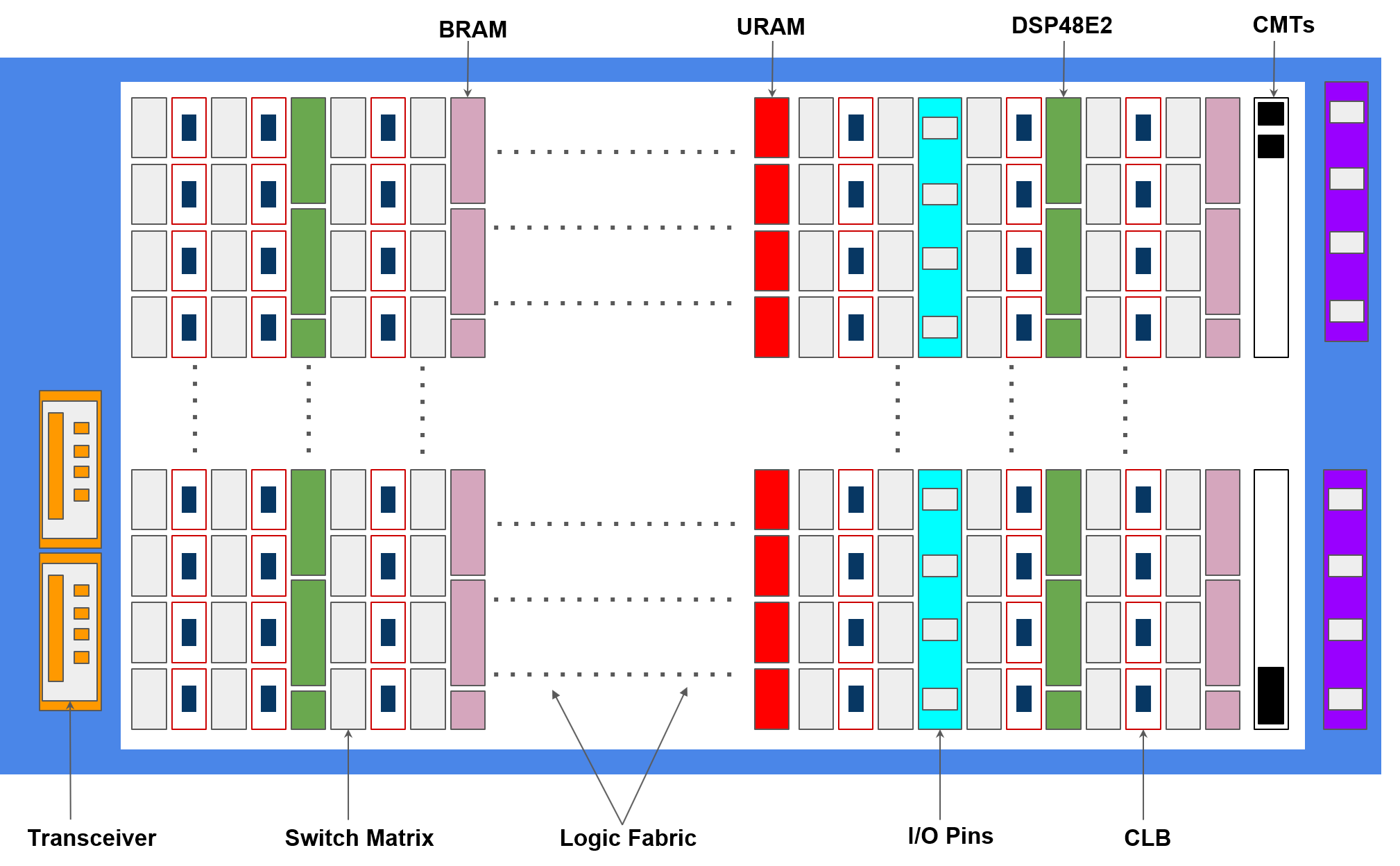}
    \caption{Example UltraScale+ layout}
    \label{fig:fpga}
\end{figure}

This thesis hopes to focus on a novel method of representing neural networks, that lends us advantages during inference. We hope to identify appropriate use cases for this method of developing neural network topologies, and gain insight into the behavior of  aggressive sparsification and quantization of neural networks along with the ramifications of design decisions to hardware cost.

\section{Field-Programmable Gated Arrays}

\marginpar{\centering
\includegraphics[width=120pt]{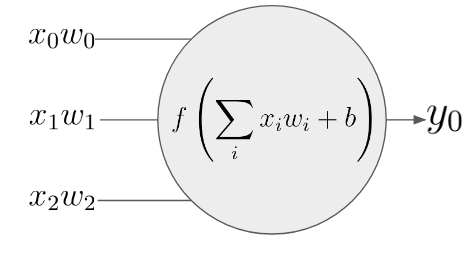}
\captionof{figure}{Illustration of a neuron.}
\label{fig:neuron}
}

\marginpar{\centering
\vspace{10pt}
\includegraphics[width=100pt]{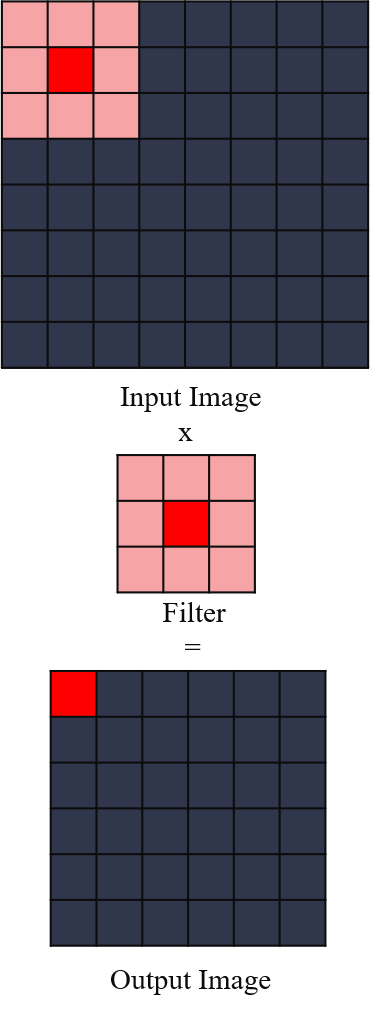}
\captionof{figure}{Illustrating Convolution.}
\label{fig:convolution}
}

The basic architecture of a Xilinx FPGA is a two-dimensional array of digital logic elements, that are grouped into Configurable Logic Blocks (CLBs). Each CLB is composed of flip-flops, Look-Up Tables (LUTs). These CLBs are connected together with programmable interconnects and switch matrices. Modern FPGAs have many improvements, one of which is larger memory blocks Block RAMs (BRAMs) and Ultra RAMs (URAMs). These larger memory elements are dense and fast, and have many applications. Modern FPGAs also have integrated arithmetic blocks (DSPs) which are capable of multiplication, addition/subtraction, and other logical functions. \\
\cref{fig:fpga} illustrates the resource layout for an UltraScale+ FPGA. At a high level of abstraction, the FPGA device has vertical regions with an array of resources. A large proportion of the FPGA fabric is dedicated to general purpose logic. CLBs (Configurable Logic Blocks) that are composed of LUTs and Flip Flops. The I/O Blocks are arranged in banks and support a wide variety of interfacing standards. The DSP48E2s are Digital Signal Processing slices that support arithmetic, as well as can be used for barrel shifting, pattern detection, and much more.\\
From \cref{st:tab:fpgaresources}, it becomes evident that we have far more LUTs than we have DSP slices. While a DSP slice $(DSP48E2)$ has functionality ranging from a 48-bit logic unit to being a $27\times18$ two's complement multiplier, our approach hopes to sparsify and quantize neural networks to an extent where it becomes more beneficial to configure these LUTs to map to neurons. \\
One of FPGA's fundamental building block is a '$K-LUT$', which can perform any $K$ input boolean operation. In this thesis, we study how niche domains can benefit vastly by viewing a neuron as an boolean function transform, which maps $X$ input bits to $Y$ output bits. This is particularly intriguing as we can convert a neuron to a configuration of such '$K-LUTs$' and achieve high throughput. \\ 

We propose utilizing the individual $K-LUTs$ to perform arbitrary non-linear functions to a fixed number of activation bits. The advantage is that we can have full-precision weights, and the result is discovering a sparse topology with activation quantization.

\begin{table}
    \begin{sidecaption}[FPGA Resource Table]{%
        Resources available in Xilinx  UltraScale\texttrademark FPGAs. Device names with a 'K' prefix refer to the Kintex\textregistered  UltraScale\texttrademark  FPGAs, names with a 'XCV' prefix refer to the Virtex\textregistered  UltraScale\texttrademark  FPGAs.
    }[st:tab:fpgaresources]
    \centering
    \begin{threeparttable}
    \renewcommand{\arraystretch}{1.2}
    \begin{tabular}{lrrrr}
        \toprule 
        Device                                   & CLB LUTs & BRAMs (18Kb) & DSP Slices  \\
        \midrule
        KU025 \                                  &  145,440  &  720      &     1,152   \\
        KU060\                                   &  331,680  & 2,160     &     2,760   \\
        XCVU065\                                 &  358,080  & 2,520     &     600   \\ 
        KU115\                                   &  663,360  & 4,320     &     5,520   \\ 
        XCVU440     \                            &  2,532,960& 5,040     &     2,880 \\ 
        \bottomrule
    \end{tabular}
    \end{threeparttable}
    \end{sidecaption}
\end{table}


\section{Deep Neural Networks}
A Deep Neural Network can be used for an array of tasks, from classification, segmentation, detection to actually generating new data after learning the semantics of the task. In our thesis, we focus on viewing each neuron as a boolean function, and limit ourselves to identification or classification tasks.

\newthoughtpar{Linear Layers}
We can think of a neuron in a linear layer as a function which applies a linear transformation on an input of dimension $I$ and results in an output of dimension $O$. We can further add a non-linearity to such a layer, like the ReLU activation function. In Figure \ref{fig:neuron}, we demonstrate a neuron of fan-in of 3, and has 1 output. From a boolean function perspective, if $x_{i}$ is the input, and $y_{0}$ is the output, and both the input and output are of 16-bit fixed point data-types, then we can expect the boolean function '$f$' to be $f : B^{48} \rightarrow B^{16}$, where $B = \{0, 1\}$. This would require a huge truth table, needing around $4.50\times10^{15}$ bits of storage. It is easy to see that such a neuron is unfeasible to implement on any FPGA. We therefore need to aggressively prune the number of inputs to such a neuron, as well as the bit-width of the input and outputs. Later in this document we will describe the relation of input and output bits of a neuron with the hardware cost as well, which will aid us in design decisions.

\newthoughtpar{Convolutional Layers}
Figure \ref{fig:convolution} represents a simple convolution operation, where the input image has 1 channel, and the filter depth is 1. It is natural to view these filters as neurons, that respond differently to different spatial features. When the filter correlates well with a region of the input image, the response in the corresponding output image is strong. Since the contents of the filter remain the same at inference, we can also represent a filter convolution with simple truth tables. Such a truth table would essentially represent the boolean function $f : B^{144} \rightarrow B^{16}$, where $B = \{0, 1\}$. This assumes that the Input and Output image pixels are all represented by 16-bit data-types. This is not a fair estimate of the dimensionality of the boolean function due to the fact that we are ignoring the number of input feature maps, which are considerably high in a typical convolutional neural network.

\subsection{Quantized Neural Networks}
Quantized Neural Networks have been a very attractive avenue for researchers in the deep learning community due to the benefits it provides for inference. Many techniques to quantize neural networks have been proposed, which broadly can be classified as deterministic and stochastic quantization. There are three components that are generally quantized. The weights, activations and the gradient. \\
We see notable inference time improvements in performance when quantizing weight and activations, while the primary benefit of quantizing gradients is to save on communication cost in distributed training. It is worth noting that it quantizing gradients is generally not advisable as high precision gradients are required to make the optimizer converge. \\
Ever since quantized neural networks have been explored, FPGA implementations of Neural Network accelerators have become increasingly better at inference with low precision bit-widths for both weights and activations. \cite{umuroglu+:FPGA2017finn} proposed BNN accelerators with a dataflow-style architecture where processing engines are instantiated for every layer. LUTs were heavily utilized to provide XNOR-Popcounts and accumulate operations and weights and intermediate activations were stored in BRAMs. This research also implemented a folding scheme to allow for larger networks to be mapped to smaller devices. \cite{wang2019lutnet} introduced LUTNet, that took a pruned version of ReBNet~\cite{ghasemzadeh2018rebnet} in which some of the XNOR-Popcount operations are mapped more effectively to $k$-input LUTs. \cite{murovivc2019massively} implemented binarized networks which have been fully unrolled and implemented directly into LUTs of a small FPGA. This unrolling allowed the synthesis optimization tool to simplify significant portions of the compute logic. \\

\subsection{Sparse Neural Networks}
In sparse neural networks, each layer of neurons receive input from only a few connections from the previous layer of activation. In typical dense layers, every neuron receives input from every activation from the previous layer. We identify three key methods to sparsify Deep Neural Networks; A Priori Fixed Sparsity, Iterative Pruning and Learned Sparsity. In A Priori Fixed Sparsity, a network is initialized with a certain connectivity pattern, which remains fixed throughout training. Two examples of this approach are Deep Expander Networks~\cite{prabhu2018deep} and RadiX-net~\cite{DBLP:journals/corr/abs-1905-00416}. Iterative pruning methods generally use the magnitude of weights to prune a dense network. Recent methods utilize other metrics for this task well. Learned Sparsity methods such as Discovering Neural Wirings~\cite{DBLP:journals/corr/abs-1906-00586} and Neuroevolution of Augmented Topologies~\cite{stanley2002evolving} are aproaches that use some form of learning to identify important connections, establishing and pruning connections as desired. 


\chapter[Hardware-Software Codesign]{Hardware/Software Codesign}\label{ch:fpga-codesign}
\openepigraph{Because of the nature of Moore's law, anything that an extremely clever graphics programmer can do at one point can be replicated by a merely competent programmer some number of years later.}{John Carmack}
The process of achieving desired functionality on a system by exploring a design space and mapping tasks to resources within system constraints is referred to as Hardware/Software Codesign~\cite{6172642}. 
\newthoughtpar{Implementing Neural Networks on FPGAs}
To aid our understanding of how a neural network is optimized for inference on an FPGA, we glance at the HLS-RFNoC workflow. This is an open source workflow for deploying a neural network for inference on an FPGA for RF signal processing~\cite{Charles2013}. This implementation uses Vivado HLS to generate custom HDL for the neural network. HLS4ML~\cite{nhan_tran_2019_3359214} is used to create the neural-network C++ code used with Vivado HLS. 
The steps can be summarized as below:
\begin{itemize}
    \item Prototype and train a neural network with industry standard machine learning framework.
    \item Use the HLS4ML C++ generation tool to create a starting point for running Xilinx HLS.
    \item Synthesize the HLS to evaluate resource usage. Once the network is implemented in fixed-point, run HLS synthesis to create HDL code.
    \item Module resulting from HDL generation is inserted into an RFNoC Compute Engine. A user testbench is then optionally created to stimulate the CE and validate its functionality. 
    \item The Compute Engine is then built into an FPGA Image using the RFNoC image building workflow.
\end{itemize}

There are some key abstractions that may make generating an efficient run-time strategy difficult. The first being, there is no insight into how a specific prototype topology would map to an FPGA in terms of cost. The entry point for most FPGA vendor tools are Register Transfer Level (RTL) designs, expressed in VERILOG or VHDL~\cite{umuroglu2018accelerating}. This brings about the second, the compiler is responsible for translating the neural network into the language of the instruction set, scheduling memory transfers on the target hardware. In the next chapter, we will further discuss how these problems manifest negatively in hardware/software codesign and discuss our proposed design methodology.

\begin{table}
    \begin{sidecaption}[Static Mapping Cost]{%
        Static Mapping Cost to 6:1 LUTs for Fan-In bits.
    }[st:tab:staticlutcost]
    \centering
    \begin{threeparttable}
    \renewcommand{\arraystretch}{1.2}
    \begin{tabular}{lrrrr}
        \toprule 
        Fan-In & Number of 6-LUTs & Truth table bits & LUT config bits & \% utilized \\  \midrule

        6 \            &  1  & 64   &     64   & 100\%    \\ 
        7\             &  3  & 128  &     192  & 66.67\%  \\ 
        8\             &  5  & 256  &     320  & 80\%     \\ 
        9\             &  11 & 512  &     704  & 72.73\%  \\ 
        10 \           &  21 & 1024 &     1344 & 76.19\%  \\ 
        11\            &  43 & 2048 &     2752 & 74.42\%  \\ 
        \bottomrule
    \end{tabular}
    \end{threeparttable}
    \end{sidecaption}
\end{table}

\section{Fan-in, Neurons and LUTs.}
One of the central concepts we must grasp to be able to delve into how we can effectively generate topologies in a Machine Learning framework which effectively map to FPGAs is how a Neuron with specific total Fan-In bits and output bits are mapped to LUTs. A brief mapping cost has been detailed in \cref{st:tab:staticlutcost}.
We deal with the mathematics behind it and attempt to give the reader a visual understanding of how the process is carried out.

\newthoughtpar{Mapping to 6:1 LUTs}
\cref{fig:71lutto61luts} and \cref{fig:81lutto61luts} depict how we can map a 7:1 and 8:1 Truth Table to 6:1 LUTs respectively. We can extend similarly to larger input-output bit configurations, which is being discussed below in more detail.

\marginpar{\centering
\includegraphics[width=130pt]{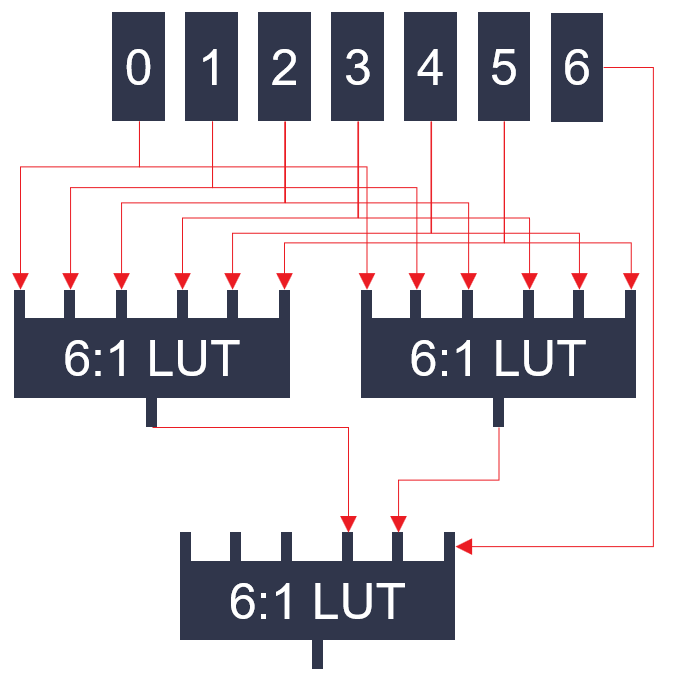}
\captionof{figure}{Mapping a 7:1 Look-up Table to 6:1 LUTs.}
\label{fig:71lutto61luts}
}
\marginpar{\centering
\includegraphics[width=130pt]{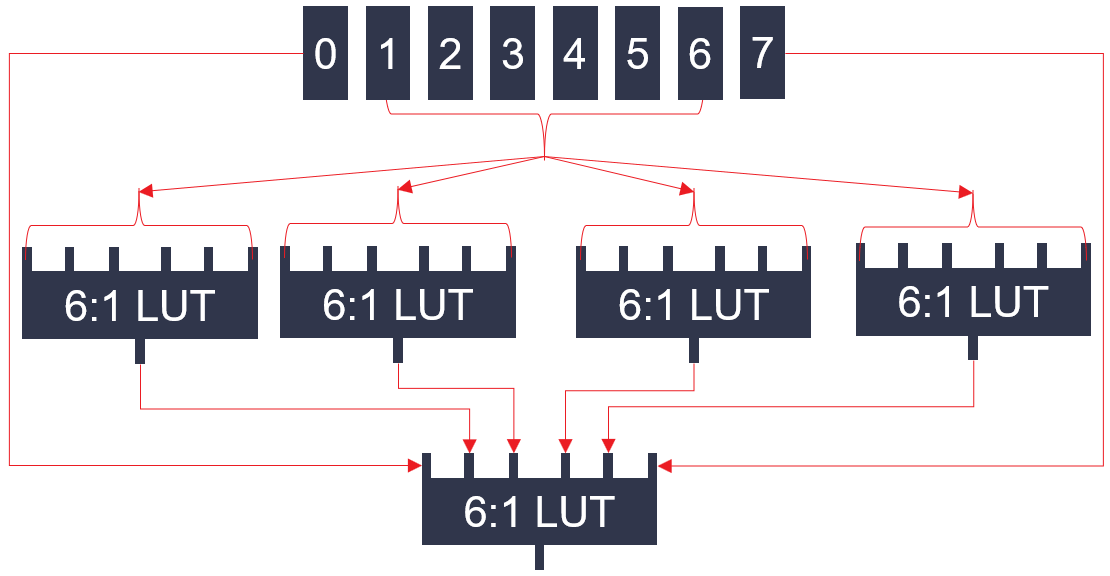}
\captionof{figure}{Mapping a 8:1 Look-up Table to 6:1 LUTs.}
\label{fig:81lutto61luts}
}

\newthoughtpar{Closed Form equation for LUT cost.}
We attempt to develop a Closed Form equation for the theoretical LUT cost of implementing a specific Neuron with arbitrary Fan-in and Fan-out.
The recursive form of the equation for the LUT cost of a neuron with $N$ fan-in bits and $M$ output bits is given by \eqref{recursiveform1}
\begin{equation}
    LUT_{N, M} = M\times(2\times(\frac{LUT_{N-1, M}}{M}) - (-1)^{N})
    \label{recursiveform1}
\end{equation}

This can be further simplified to \eqref{recursiveform2}. 
\begin{equation}
    LUT_{N, M} = M\times(\frac{LUT_{N-2, M}}{M} + 2^{N-6})
    \label{recursiveform2}
\end{equation}

Which in its closed form gives us \eqref{lutcostcloseform}. Note that these LUT costs will only be valid for hardware building blocks composed solely from 6:1 LUTs. Getting the minimum LUT cost from the combination of the wide array of memory types that an FPGA has (5:2 LUTs, 6:1 LUTs, different BRAM configurations) is a design space exploration problem. This elementary equation is integrated into the design flow to give the user a rough hardware cost heuristic. There are many ways to decrease this cost, which will be discussed later.
\begin{equation}
    LUT_{N, M} = M\times(\frac{2^{N-4} - (-1)^{N}}{3})
    \label{lutcostcloseform}
\end{equation}

\chapter{Mapping Neurons to Hardware}\label{ch:mapping}
\marginpar{\vspace{-10pt}\includegraphics[width=45mm]{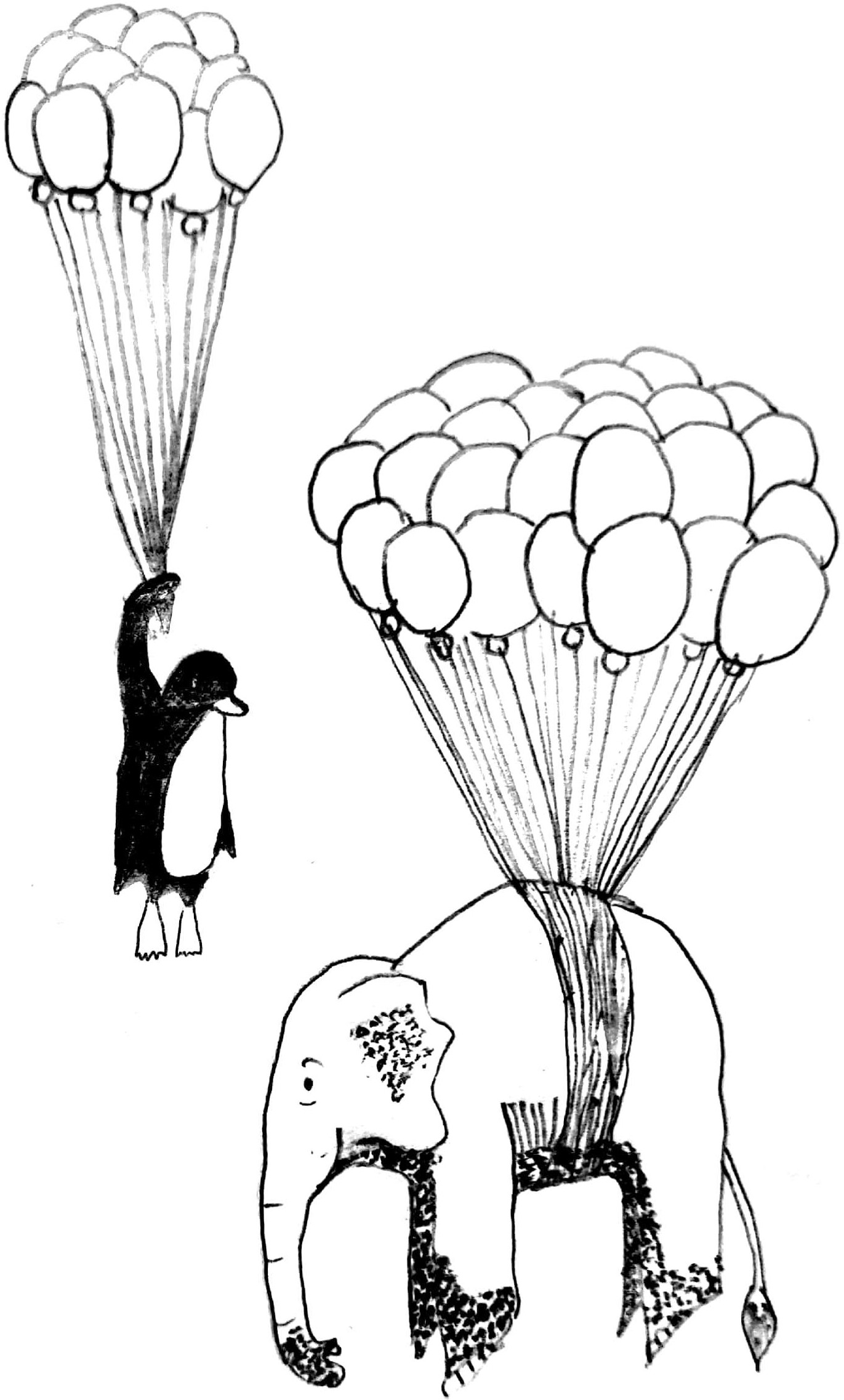}\\\footnotesize\hphantom{.}\hfill---Sneha Khandelwal}

Prototyping a neural network on a target hardware has a recurring cost, as often an engineer has no true estimate of the resources that will be required. This problem is further aggravated by the complex interaction of a topology and the underlying hardware, a compiler which translates the neural network to the target hardware typically yields inefficient deployment strategies.

In this thesis, we present a novel method of mapping neurons to hardware. This is done without the need for a custom accelerator architecture or a scheduler. If we turn back to our initial representation of a neuron as a boolean function $f: B^{ip} \rightarrow B^{op}$ where $B = \{0, 1\}$, we can hope to represent such a neuron with a look up table which requires $2^{ip} \times (op + ip)$ bits. Here, it becomes evident that in topology design, our most crucial design consideration should be to minimize $ip$, which is the number of input bits a neuron takes. However, as a neural network is essentially several layers of neurons placed sequentially, an implicit bound is also imposed upon $op$. While this may sound discouraging, a neural network is often heavily over-parametrized. The paper Deep Compression~\cite{han2015deep} used pruning, trained quantization and huffman coding to reduce the storage requirements of their neural networks by about $40\times$ without affecting their performance. 
\newthoughtpar{Prior works in accelerating neural networks}
There have been aggressive efforts in quantizing neural networks. Ever since Binary Neural Networks~\cite{courbariaux2016binarized} delivered respectable performance, FPGA implementations of quantized neural networks have been booming as well. \cite{umuroglu+:FPGA2017finn} proposed BNN accelerators with dataflow-style architectures (processing enginers instantiated at every layer). \cite{umuroglu+:FPGA2017finn} also implemented a folding scheme for larger neural networks so that they can be mapped to smaller devices. LUTs are heavily utilized in this implementation. \cite{wang2019lutnet} introduced LUTNet, which is a LUT optimized compression scheme which achieves very high LUT density. \cite{abdelsalam2018polybinn} introduced POLYBiNN, which utilized 110k 6-input LUTs to achieve an accuracy of $97.18\%$ on MNIST. POLYBiNN achieved 100M FPS on MNIST.
\section{Sparsity and Quantization}
\subsection{Sparsity}
In sparse neural networks, each neuron receives inputs from only a few connections of the previous layer. Often, it is beneficial to have regular sparsity, where pruning is done in a manner where neurons without value are pruned completely. This is not true for an LogicNet. We need to sparsify a network in a manner that every neuron maintains a small number of connections. This requirement has been demonstrated in \cref{fig:sparsify}. We identify three methods of Neural Network pruning: \textit{A-Priori Fixed Sparsity, Iterative Pruning and Learned Sparsity}. 
\marginpar{\centering
\includegraphics[width=120pt]{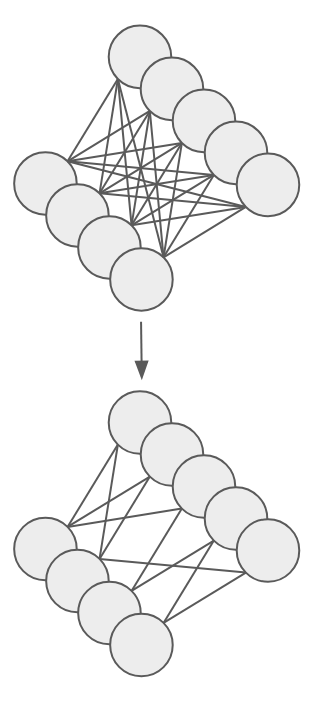}
\captionof{figure}{Sparsification per neuron.}
\label{fig:sparsify}
}
\newthoughtpar{A-Priori Fixed Sparsity}
We adapt concepts developed in Deep Expander Graphs~\cite{prabhu2018deep}. We use the concept of Random Expanders to explain our sparsity requirements. \\
\textbf{Random Expanders:} A random bipartite expander of degree $D$ on two Vertex Sets $U, V$ is a graph in which for every vertex $v \in V$, the $D$ neighbours are chosen independently and randomly from $U$.\\
We want to construct a neural network which corresponds to a sparse graph, where each vertex has $D$ neighbours. To extend this concept to convolutions, we need to view each set of convolutional kernels as one neuron, or as a vertex. This implies that for a convolutional filter set $(1, C, H, W)$, we can only have $D$ non-zero weights. 

\marginpar{\centering
\includegraphics[width=120pt]{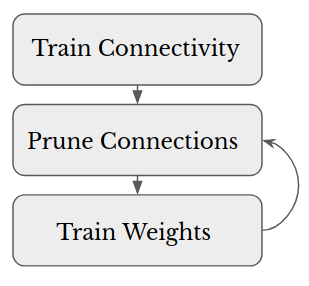}
\captionof{figure}{Training Pipeline.}
\label{fig:iterativepruning}
}
\newthoughtpar{Iterative Pruning}
Learning the right connections is an iterative process, the pruning is done by zero-ing the weights with the smallest magnitude. Pruning followed by retraining is one iteration, and per neuron pruning decay rates are calculated such that the number of weights zeroed out by the end of training gives rise to a fixed sparsity, where every neuron has $D$ neighbours. The training pipeline is presented in \cref{fig:iterativepruning}. Each iteration is greedy, and unlike the original implementation, we focus on pruning weights inside each neuron. This gives us a strict control over the connectivity of each neuron, consequently helping us in controlling the size of the truth table of the equivalent boolean function of the neuron. 

\newthoughtpar{Sparse Learning}
In the paper \textit{Sparse Networks from Scratch: Faster Training without Losing Performance}~\cite{dettmers2019sparse}, there are three steps; (a) Re-distribution of weights, (b) Pruning of weights, (c) Re-growing weights. To explain further, we take a simple $MLP$ model and index the layers as $l$, neurons as $n$ and synapses as $s$. The authors maintain an 'exponentially smoothed gradient' for every weight in the MLP. This can be described by the equation below.
$$
M_{l, n, s}^{t+1} = \alpha M_{l, n, s}^{t} + (1 - \alpha) \frac{\partial E}{\partial W_{l, n, s}}
$$
\newthoughtpar{Re-distribution of weights}
The normalized mean of the element-wise momentum for all non-zero weights in each layer is taken, as described below.
$$
M^{mean}_{l} = \frac{\sum_{n}\sum_{s} M_{l, n, s}}{\sum_{l}\sum_{n}\sum_{s} M_{l, n, s}}
$$
The resulting proportions are the momentum magnitude contributions for each layer. This is then used to calculate the number of weights to be regrown. For a network with sparsity $r$, the Regrowth for each layer is calculated below.
$$
Regrowth_{l} = n(Params)\times(1-r)\times M^{mean}_{l}
$$
\textbf{Pruning}\\
The authors prune a proportion $p$ (prune rate) of the weights with the lowest magnitude in each layer.\\
\textbf{Regrowth}\\
Weights are re-grown by enabling gradient flow of zero-valued weights which have the largest momentum magnitude.
\begin{figure}
    \centering
    \includegraphics[width=450pt]{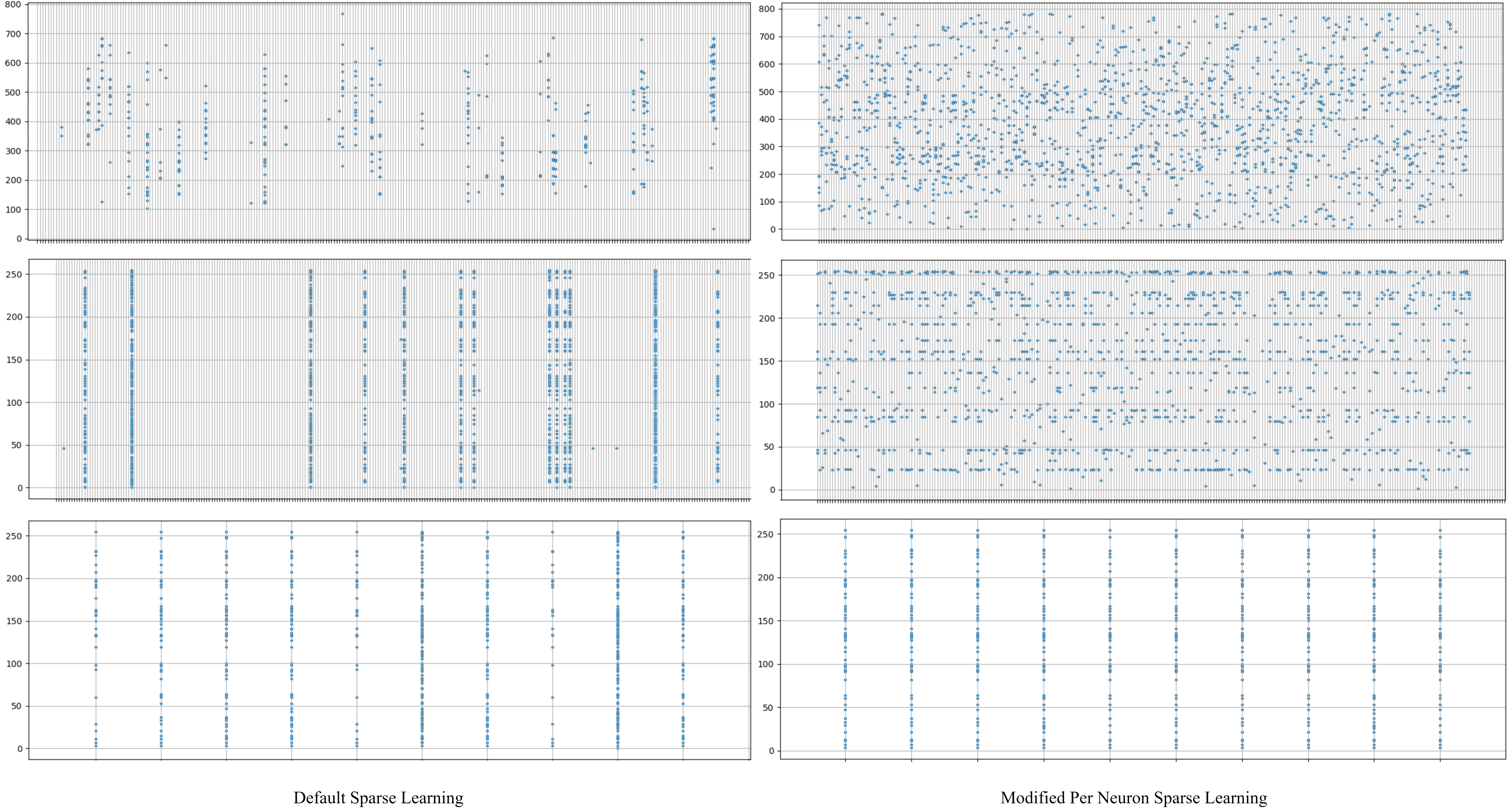}
    \caption{This figure demonstrates the per neuron connectivity for a 3-Layer Neural Network.}
    \label{neuronsparsity}
\end{figure}
\newthoughtpar{Modifying Sparse Momentum Pruning}
As detailed in the \cref{algorithm1}, we had to make some notable changes to the Sparse Learning Algorithm. \cref{neuronsparsity} Attempts to measure the per-neuron fan-in for each neuron. Each divison in the X-Axis represents a Neuron, and each divison in the Y-Axis represents an incoming activation pixel. The dots represent which activation pixels each neuron is connected to. In the default sparse learning algorithm shown in \cref{neuronsparsity}, we can see that a few neurons have very high fan-in, while the others are pruned away. We desire a low, uniform per neuron fan in, which is achieved in the Modified Per Neuron Sparse Learning algorithm. 


\begin{algorithm}
\caption{A Pruning Step for a L-Layer MLP}
 \hspace*{\algorithmicindent} \textbf{Input:} Layer $i$ to $l$ with: Momentum $M_{i}$; Weight $W_{i}$; binary $Mask_{i}$; prune rate $p$; Neuron Fan-in $F$; Neuron Prune $P1$; Neuron ReGrowth $R1$;\\
 \hspace*{\algorithmicindent} \textbf{Output:} Updated weight $W^{t+1}$; TotalMomentum; TotalNonZero;
\begin{algorithmic}[1]
\For{$i$ = 1 to $l$}
    \State MeanMomentum$_{i}$ $\leftarrow$ MeanMomentum$_{i}$ + $mean(abs(M_{i}[W_{i}\neq0]))$
    \State TotalMomentum $\leftarrow$ TotalMomentum + MeanMomentum$_{i}$
    \State NonZero$_{i}$ = $\sum(W_{i}\neq0)$
    \State TotalNonZero $\leftarrow$ TotalNonZero + NonZero$_{i}$
\EndFor
\For{$i$ = 1 to $l$}
    \For{Neuron in Layer $i$}
        \State Prune($P1$, Neuron)
        \State ReGrow($R1$, Neuron)
\EndFor
\EndFor
\end{algorithmic}
\label{algorithm1}
\end{algorithm}

\newthoughtpar{Discussion}
Our modification of the Sparse Learning Algorithm has some shortcomings. Due to the strict requirement of a fixed per-neuron Fan-in, the $MeanMomentum_{i}$ has no redistribution utility in our implementation. This implementation simply prunes weights per neuron by magnitude, and re-grows weights per neuron by their momentum. It is important to note that, in the future it may be a worthwhile endeavor to create a better re-distribution strategy which allows for variable fan-in per neuron based on the MeanMomentum per neuron. In our testing, such a re-distribution strategy had adverse effect on hardware cost, with very little to no gain in accuracy. We leave these loose ends as part of the algorithm so that the reader is aware of all the variables that are still being tracked in our implementation.

\subsection{Quantization}
To support quantization, we utilize the Machine Learning PyTorch Library Brevitas~\cite{alessandro_pappalardo_2019_3525102}. Brevitas implements a set of building blocks to model a reduced precision hardware data-path at train time. Our methodology of design places no constraint on the weight data-type. We however, do have a constraint on the activation bit-width. We thus utilize the \textit{QuantReLU} and \textit{QuantHardTanh} layers to support Binary and uniform integer quantization of activations.
In a quantization flow with integer uniform quantization for activations, the bit-width together with the sign determines the min and max integer values used for scaling and clamping. Since the range of QuantReLU output is $QuantReLU(input) \in \mathbb{R}^{+}$, we do not have to worry about the sign bit for scaling. The \textit{QuantReLU} and \textit{QuantHardTanh} return a QuantTensor (NamedTuple), which propagate the following information:
\begin{itemize}
    \item quant\_tensor: The quantized tensor in dequantized representation (floating-point order of magnitude)
    \item scale\_factor: The scale factor of quant\_tensor
    \item bit\_width: The precision of quant\_tensor in bits.
\end{itemize}

\section{Quantifying Hardware Costs}
Now that we have discussed the ways we have tried to enforce sparsity and the quantization scheme being used, it is important to understand how sparsity and quantization work together to reduce the hardware cost for a neuron. We will deal with the implications of different sparsification and quantization methods on accuracy when we discuss the data-sets LogicNet was tested on.

\section{Research Recess  I -- Sparsity}

In the previous sections, we focused on methods to port neural networks of sizeable dimensions to the LogicNet topology. In spirit of research, a discourse on why some methods work, some don't could be of great value. 

\subsection{What are some methods to distribute non-zero weights?}
At first, this question may seem to have been answered already. While that is true to some extent, it is worth formalizing. The utility of exploring re-distribution methods is to find natural ways to make neurons adhere to specific fan-in constraints. In our A-Priori Fixed Sparsity methods, we have no focus on neuron importance. Our modified implementation of Sparse Learning algorithm simply clips off the same proportion of weights from each neuron. This may be very unnatural. We refer to the work by \cite{evci2019rigging} and consider two weight distribution strategies. The original work also considers an {Erd\H{o}s-R\'{e}nyi-Kernel} method, which is an extension of {Erd\H{o}s-R\'{e}nyi} method which takes the kernel size into consideration. 
\begin{itemize}
    \item \textit{Uniform:} The Sparsity of each layer is the same as the Total Sparsity.
    \item \textit{Erd\H{o}s-R\'{e}nyi:} The sparsity of each layer scales with 1 - $\frac{n^{l-1} + n{l}}{n^{l-1}\times n^{l}}$ where $n^{l}$ denotes the number of neurons at layer l. This method gives higher sparsity to layers with more parameters and lower sparsities to smaller ones.
\end{itemize}

\subsection{Ensembling and the {Erd\H{o}s-R\'{e}nyi} Method}
Ensembling has shown great benefits in our experiments, thus a way to take sparsity distribution in account and adjusting our model layer costs accordingly is an interesting exercise. 
We can benefit from the {Erd\H{o}s-R\'{e}nyi} method of weight redistribution, as it naturally aligns with our view of neural networks and allows them to scale a bit better. As the complexity of the LUT cost of a layer is$O(2^{B})$ (B is the fan-in in bits), if a layer with $N1$ neurons is made more sparse than a layer with $N2$ neurons where $N1>N2$. The LUT cost of the neurons become proportional to $2^{B1}{N1}$ and $2^{B2}{N2}$, note that as the sparsity of the larger layer is greater $\mathbb{E}(B2)>\mathbb{E}(B1)$. We thus allow ourselves to ensemble many of those smaller layers or the larger layers and balance the cost. The expression for the number of such smaller layers can be 'ensembled' can be provided by simply dividing the two LUT costs as to obtain $N2\frac{2^{B2-B1}}{N1}$. The inverse is true if the larger layer requires lesser LUTs. \\
We can also try adjusting the {Erd\H{o}s-R\'{e}nyi} method to take the exponential nature of the cost of a layer into account. To take this idea further, we do not need to proportionally ensemble intermediate layers, rather we can ensemble intermediate layers of arbitrary sizes as long as it respects the above proportionality in cost. Which in itself is only in place to have balanced sparsity between layers. What utility redistributing weights to balance sparsity across layers provides is an interesting research question.\\

\subsection{Is Sparsification Neural Architecture Search?}
If we think more closely about what we are doing, it becomes evident that for all our models, the maximum size of sparse models is bounded by the largest dense model we start with. This was pointed out by~\cite{evci2019rigging} and while it may seem trivial, it implies this statement not only from a parameter masking perspective but more interestingly, from a topological perspective. Some methods of Sparsification can be 'tagged' as Neural Architecture Search. It may make sense to look more closely at more 'esoteric' methodologies of discovering topologies which are sparse. We may tailor topologies ourselves, as we attempted in the previous subsction on \textit{Ensembling and the {Erd\H{o}s-R\'{e}nyi} Method}, or we can look at recent research in the field of discovering neural connections. Some of which are ~\cite{Wortsman2019DiscoveringNW} ~\cite{Mocanu_2018} ~\cite{liu2019sparse} ~\cite{bellec2017deep}. \\

The Sparse Evolutionary Training algorithm essentially replaces fully connected layers with Sparse Connected layers having the {Erd\H{o}s-R\'{e}nyi} topology. For each training epoch, they take each bipartite Sparse Connected layer, removing a fraction of the smallest positive and largest negative weights. They also randomly add new connections in the same amount as previously removed. \\
We need an algorithm that not only converts a fully connected layer to a Sparse Bipartite Graph, but treats each neuron with certain fan-in as nodes in a graph with arbitrary connectivity (without hierarchy). Ideally, we would like to implement a neural architecture search which does not pay heed to the hierarchial structure of DNNs. The search space for such a task is too huge. Sparsification of each neuron leads to a significantly smaller search space, but the order of complexity remains the same. \\
One of the most interesting papers which support such an idea is by \cite{Wortsman2019DiscoveringNW}. In this paper, a set of real edges $\eta$ is considered, with a set of \textit{hallucinated} edges $\xi_{hal} = V\times V \textbackslash \xi$. The magnitude of a weight is strengthened if the gradient pushes the activation in alignment with a certain edge. Over iterations, if it is strengthened enough it can join the set of real edges from the hallucinated edges. \\
It is extremely important to conduct this 'neural architecture search' in a manner such that the search space is navigated effectively. If the neural network has no 'hierarchy', it becomes a much harder task. Trying to discover sparse topologies which are not hierarchical in nature is one of the most exciting future areas of work for this thesis. 

\begin{fullwidth}
\part{LogicNet: A Library for Mapping HBBs to NEQs}
\end{fullwidth}
\chapter{The LogicNet Design Flow and Library}\label{ch:intro-fpganet}
\openepigraph{\mbox{Premature optimization is the root of all evil}}{Donald Knuth}

\newthought{Synopsis}\hspace{1.5em}
This and the next chapter that follows hopes to delve into the specifics of the LogicNet design flow and the library. This chapter discusses how we can train efficient DNNs that map directly to FPGAs along with the layer types we support in this library, briefly touching the functionality provided. The next chapter focuses on the functions supported to aid the design automation process. The basic LogicNet Design Flow:
\begin{figure}
    \centering
    \includegraphics[width=380pt]{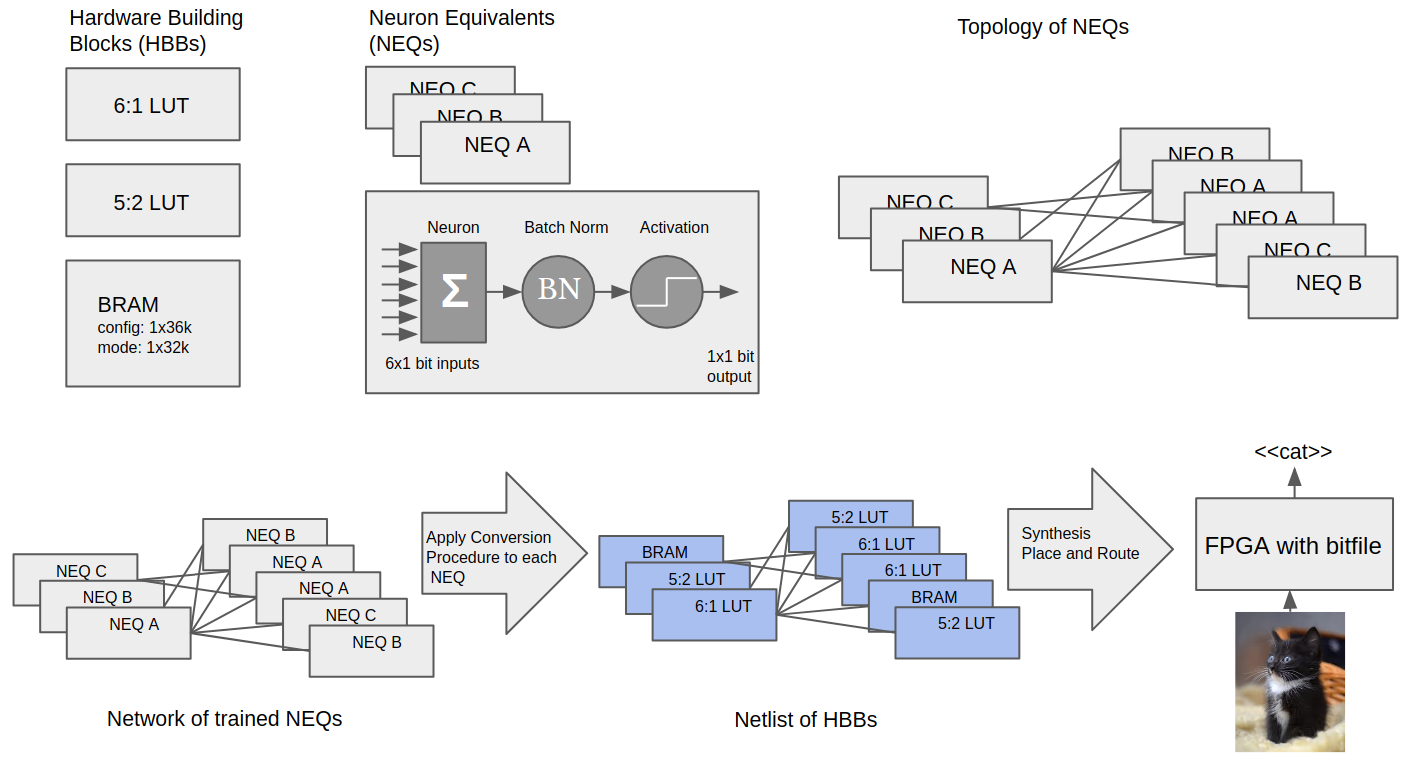}
    \caption{The Design Flow for LogicNet}
    \label{fig:designflowfpganet}
\end{figure}
\begin{itemize}
    \item Define Hardware Building Blocks (HBBs), Neuron Equivalents (NEQs) and their conversion procedure.
    \item Define a Deep Neural Network in the LogicNet library and train.
    \item Convert trained network of NEQs into a netlist of HBBs using the conversion procedure. (All steps of the conversion procedure are supported by the LogicNet library itself).
    \item Postprocess the netlist and synthesize to obtain a bitfile.\\
\end{itemize} 
In LogicNet, every layer type has an implicit input Quantizer. The reason for this design choice is that input quantization is one of the most crucial aspects of an LogicNet, and output quantization is optional. The LUT cost increases linearly with the output bit-width of a neuron, whereas increases exponentially with the input fan-in bits. We lend the user some flexibility in output quantization, as it may be particularly important for the final layer. 
\section{Quantizer}
The Quantizer is a relatively simple module, which uses the Brevitas implementation of QuantHardTanh and QuantReLU. We can see clearly in \cref{fig:qhtanhqrelu} how QuantHardTanh and QuantReLU operate. The scaling factors are ignored so that we can discuss the bit-width with more ease.
If a bit-width of 1 is provided to the Quantizer, the class constructor initializes QuantHardTanh. If we provide a greater bit-width, we use a Quantized ReLU. For instance, if a bit-width of $3$ is provided, we can see that the QuantReLU will quantize any input to integer outputs from $0$ to $8$. This would also be then multiplied by a scaling factor. 
    
\marginpar{\centering
\includegraphics[width=150pt]{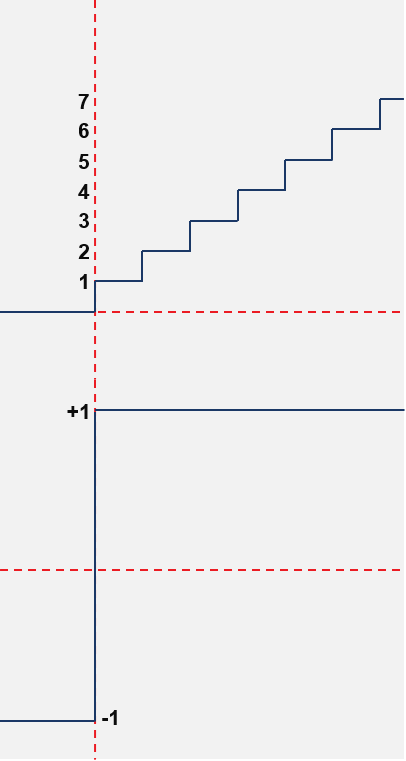}
\captionof{figure}{QuantReLU and QuantHardTanh (1 bit) Respectively. Scaling Factors not depicted.}
\label{fig:qhtanhqrelu}
}

The Quantizer returns a NamedTuple, which contains the Quantized Tensor in De-Quantized representation, with the scale-factor and the bit-width. Code Listing \ref{quantizeeg} depicts how a Quantizer operates on an input tensor.\\\\
\begin{lstlisting}[language=Python, caption=Example of a Quantizer, label=quantizeeg]
import torch
from fpganet.quantizer import Quantizer
quantizer = Quantizer(bit_width = 1, max_val = 1.61)
print(quantizer(torch.randn(10)))
>>> QuantTensor(tensor=tensor([-1.6100, -1.6100, -1.6100,  
                1.6100, -1.6100, -1.6100, -1.6100,  1.6100,
                1.6100,  1.6100],
         grad_fn=<DifferentiableGraphBackward>),    
         scale=tensor(1.6100, grad_fn=<ClampMinBackward>), 
         bit_width=tensor(1.))
\end{lstlisting}

\section{SparseLinear}
The SparseLinear layer has an input Quantizer followed by a Linear Layer with a specific per-neuron fan in and Batch Normalization. The per-neuron fan-in is represented by a layer mask, and is initialized randomly when the layer is instantiated. It is important to note that in this implementation, the random sparsity is static and not learnable. For correct Truth Table generation, it expects us to give the next module in the forward pass. This is necessary to get the output bits a neuron produces for a specific input.

\newthoughtpar{LUTs}
The SparseLinear layer has a 'LUTS' attribute, that calculates the total number of LUTs that would be required when the class is instantiated. This functionality was added to aid design space exploration. The SparseLinear layer also has a method getLUTs() which returns the LUT cost of that layer.

\newthoughtpar{Truth Table Generation}
The SparseLinear layer has a generateTruthTable() method. Once called, it assigns an OrderedDict to the truthtable attribute of SparseLinear. This OrderedDict contains the TruthTable for each neuron.

\newthoughtpar{Truth Table functional verification}
The SparseLinear layer also supports forward pass which utilizes the TruthTable. This is important from a functional verification standpoint. The forward pass of the SparseLinear module has an argument 'use\_table', which when set allows forward pass through that layer using the truthtable dictionary.

\section{DenseQuantLinear}
The DenseQuantLinear layer has an Input Quantizer followed by a QuantLinear layer and Batch Normalization. 
\newthoughtpar{LUTs}
The DenseQuantLinear layer has a 'LUTS' attribute as well. The functionality is similar to SparseLinear, but the equation used is Equation \eqref{denseqlinlutformula}. In this formula, $n(O)$ and $n(I)$ refer to the output feature count and input feature count respectively. $BW_{in}$ and $BW_{wt}$ refer to the input bit-width and weight bit-width respectively. 
\begin{equation}
    LUTS = n(O)\times(n(I)\times BW_{in}\times BW_{wt}\times 1.0699 + 10.779)
    \label{denseqlinlutformula}
\end{equation}

\section{SparseConv}
Implementing Convolutions with a Sparsity that LogicNet can leverage is a tricky task. This is due to the fact that Convolutions typically have kernels with many channels, that are not independent. The LUT cost for a fully unfolded dense convolution is shown in Equation \eqref{lutcosteqn}. Where $outpix$ refers to the number of output pixels, $oBits$ refer to output bits, $n(OFM)$ refer to the number of output feature maps, $n(IFM)$ refers to the number of Input Feature Maps, $k$ refers to the kernel size (assuming square kernels) and $iBits$ refers to the number of Input bits.
\begin{equation}
    LUTs = outpix\times oBits \times n(OFM) * LUTCost(n(IFM)*k^{2}*iBits)
    \label{lutcosteqn}
\end{equation}

\begin{figure}[h]
    \centering
    \includegraphics[width=300pt]{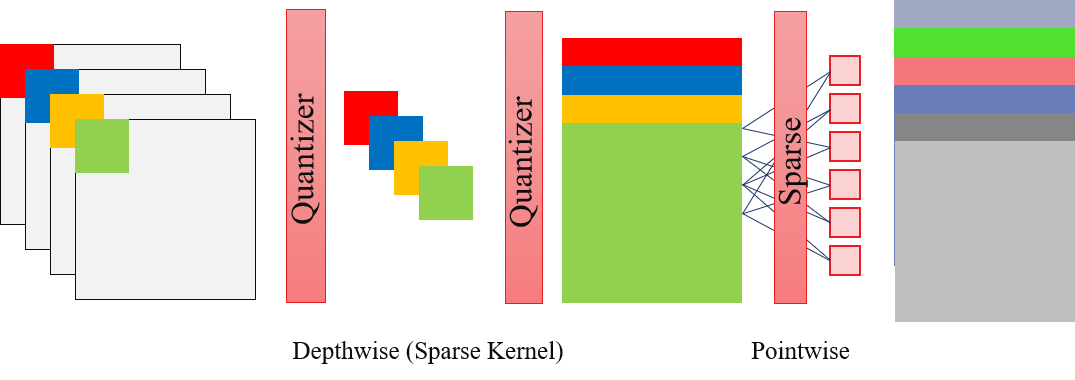}
    \caption{Depthwise Separable Convolutions in LogicNet.}
    \label{fig:dwsepconv}
\end{figure}

The most effective way we could find to deal with this is to introduce 'Depthwise Separable Convolutions' with Input and Intermediate Quantizers and Sparsity into the LogicNet library. \cref{fig:dwsepconv} depicts how such a Convolution would be implemented in the LogicNet library.\\

We give the user the ability to control the input bit-width, intermediate bit-width, as well as the sparsity of the Depthwise Kernels and Pointwise Kernels. The SparseConv also has Batch Normalization in both stages.
If the first\_layer argument is set true for SparseConv, it checks if the input channel count is 1. If that is true, it sets the Kernel count of Depthwise stage equal to the number of Output Feature Maps. This is useful because as the kernels are very sparse, it becomes harder to extract useful information out of the input image with just 1 randomly initialized sparse 2D kernel.

\newthoughtpar{LUTs}
By virtue of using Depthwise-Separable Convolution, our LUT cost is divided to two parts. The Depthwise and Pointwise LUT cost.
Depthwise LUT Cost is given by Equation \eqref{dwLUTcost}. Here, $X_{k}$ refers to the Kernel Sparsity.
\begin{equation}
    LUTs_{dw} = outpix\times obits\times n(OFM)\times LUTcost(X_{k}\times ibits)
    \label{dwLUTcost}
\end{equation}
Pointwise LUT Cost is given by Equation \eqref{ptLUTcost}. Here, $X_{s}$ refers to the Pointwise Sparsity.
\begin{equation}
    LUTs_{pt} = outpix\times obits\times n(OFM)\times LUTcost(X_{s}\times ibits)
    \label{ptLUTcost}
\end{equation}

As this is a fully unfolded convolution, the LUT cost calculation requires the $outpix$ and $n(OFM)$. Thus, the LUTS attribute is populated only after a forward pass is done on the Model.
\newthoughtpar{Truth Table Generation}
The SparseConv layer has a generateTruthTable() method. Once called, it assigns an OrderedDict to the truthtable attribute of SparseConv. The OrderedDict contains separate OrderedDict for the Depthwise and Pointwise stages, with proper indexing for the neurons. 

\newthoughtpar{Truth Table Functional Verification}
The SparseConv layer also supports forward pass which utilizes the TruthTable. It is similar in nature to the SparseLinear. It is important to note that this is purely for Functional Verification, and is thus quite slow. It loops iteratively through every input window and uses the truth table for the forward pass. 

\begin{lstlisting}[language=Python, caption=Implementing Truth Table Functional Verification, label=convFuncVerif]
rlen = int(1 + (x.shape[-2] - self.kernel_size)/self.stride)
clen = int(1 + (x.shape[-1] - self.kernel_size)/self.stride)
dw_out = torch.zeros(x.shape[0], self.inWCout if (self.first_layer and self.inWCin==1) else self.inWCin, rlen, clen)        
for batch in range(x.shape[0]):
    for h in range(rlen):
        for w in range(clen):
            vert_start = h*self.stride
            vert_end = h*self.stride + self.kernel_size
            horiz_start = w*self.stride
            horiz_end = w*self.stride + self.kernel_size
            submatrix = x[batch, :, vert_start:vert_end, horiz_start:horiz_end]
            dw_out[batch, :, h, w] = lookupConvDW(self, self.truthtable['dw'], submatrix, (self.first_layer and self.inWCin==1))
\end{lstlisting}

\chapter{Design Automation}\label{ch:designautomate}
\openepigraph{%
    Nicht Kunst und Wissenschaft allein,\\Geduld will bei dem Werke sein.
}{Johann Wolfgang von Goethe}
\openepigraph{During human progress, every science is evolved out of its corresponding art.}{Herbert Spencer}
\section{Truth Table Generation}
After the network has been trained, we need to generate the truth table for every neuron. Currently, we support a $generateTruthTable$ method with the class, to allow the user to generate truth tables for only specific layers for inspection. In practice this is specially useful, as the model size increase. \\
The truth table for a neuron with a fan-in of 6 bits requires $2^{6} = 64$ entries. However, in complicated data-sets it is impractical to keep the fan-in at only 6 bits. If we increase the number of bits for the fan-in of each neuron, the number of entries grow exponentially. For instance if the neuron has 20 input bits, it will have $2^{20} = 1048576$ entries. Clearly, for such models we not only need to allow on the go calculation of the truth table for each layer, but also the truth table for each neuron. This will allow us to extract more parallelism out of the process. Support for per-neuron calculation of truth tables will be added soon.
\newthoughtpar{Structure of Truth Table}
We can see in Listing \ref{layerttablestruct}, the structure of the truth table of a model which has 3 neurons fan-in of 3, with a bit-width of 1. The keys are the neuron IDs of the layer (in this case linear1). Once the key is accessed, we get a list of lists, the first list containing the binary input to a neuron, and the second containing the output for that specific input. While the first list does not have any utility as the input bits are inumerated in a fixed fashion, we will keep this functionality for now.

\begin{lstlisting}[language=Python, caption=Structure of the Truth Table for a layer, label=layerttablestruct]
>>> print(model.linear1.truthtable)
OrderedDict([('0',
              [['000', '001', '010', '011', '100', '101', '110', '111'],
               ['1', '1', '1', '0', '1', '0', '0', '0']]),
             ('1',
              [['000', '001', '010', '011', '100', '101', '110', '111'],
               ['1', '0', '1', '0', '1', '0', '1', '0']]),
             ('2',
              [['000', '001', '010', '011', '100', '101', '110', '111'],
               ['1', '0', '1', '0', '1', '0', '1', '0']])])
\end{lstlisting}

\newpage

\marginpar{\centering
\includegraphics[width=500pt]{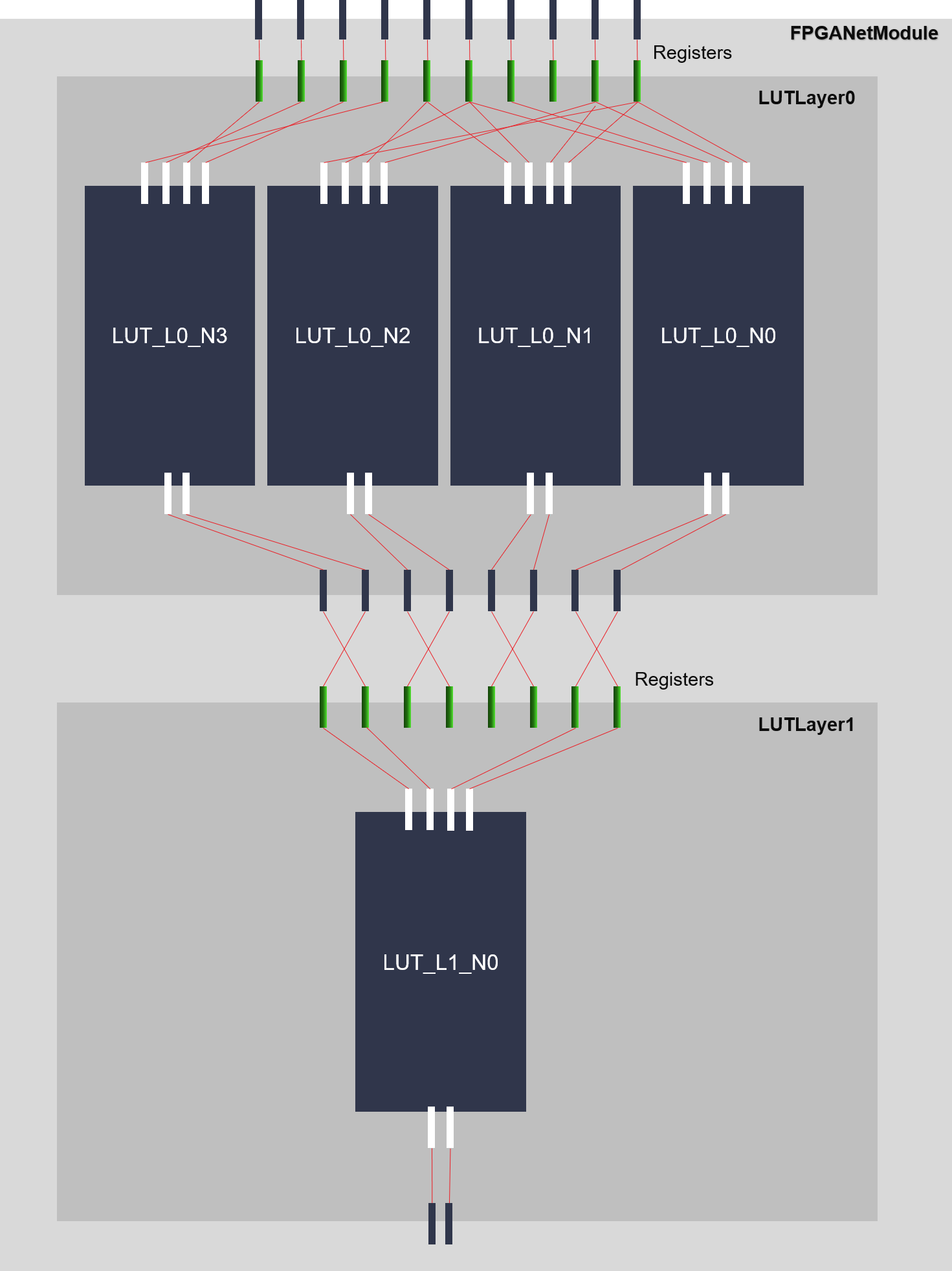}
\captionof{figure}{VERILOG Code Generation Sub-Modules.}
\label{fig:fpganetmodule}
}   

\newpage
 
\begin{table}
    \begin{sidecaption}[Size and Time estimates for VERILOG Generation]{%
        Rough estimates of the file size to store the truth table for 1 neuron with a specific fan-in in bits, and the time it takes to write the '.v' file.
    }[designautomate:tab:filesize]
\begin{threeparttable}
\begin{tabular}{lrr}
\hline
Bits & File Size(MB) & Time (seconds) \\ \hline
15   & 0.85          & 56      \\
16   & 1.8           & 119     \\
18   & 6.9           & 502     \\
20   & 29.0          & 2022    \\ \hline
\end{tabular}
\end{threeparttable}
\end{sidecaption}
\end{table}

\section{VERILOG Code Generation}

The VERILOG generator has been integrated as a VerilogGenerator class. This class takes in a model, identifies all the layers that are 'SparseLinear'. While Truth Table Generation is supported for all topologies (Linear and Convolution), we do not currently support VERILOG Code Generation for 'DenseQuantLinear' and 'SparseConv'. The SparseConv implementation will need a lot of thought, as it needs to incorporate any form of folding that may be necessary. VERILOG Generator for a Fully-Unfolded SparseConv will be integrated into the library at a later time.\\
From the code sample below, it becomes evident that we do not use any LUT primitives when generating the VERILOG code for the network. We instead define the entire truth table, and leave it up to the Logic Synthesis tool to generate the optimal 'hardware building block'. It is a very interesting research question, to explore whether a more optimal solution is discovered by the synthesis tool or does it have any benefit to define primitives ourselves. 
Further, due to the nature of how the verilog code itself is generated, the 'file size' and the time to generate the truth table explodes exponentially as well (\cref{designautomate:tab:filesize}). While the current VERILOG generator is purely sequential, we hope to extend functionality to generate the codes in a more parallel fashion. This would allow us to test a wider array of topologies faster.

\newthoughtpar{VERILOG code sample}
We attach a code sample for VERILOG code generated for the Neural Network in the previos section (Single Layer, 3 Neurons with each neuron having a fan in of 3 bits). Note that this code sample is for synthesis of a \textbf{purely combinational circuit}, no registers have been placed at the input or output. This library supports VERILOG code generation with intermediate registers, and will be discussed later. \cref{fig:fpganetmodule} clearly shows how registers are implemented at the input and the intermediate layer.

\begin{lstlisting}[language=Verilog, caption=LogicNetModule.v, label=LogicNetModulev]
module LogicNetModule (input [4:0] M0, output[2:0] M1);
        LUTLayer0  LUTLayer0_inst (.M0(M0), .M1(M1));
endmodule
\end{lstlisting}

\begin{lstlisting}[language=Verilog, caption=LUTLayer0.v, label=LUTLayer0v]
module LUTLayer0 (input [4:0] M0, output [2:0] M1
);
wire [2:0] inpWire0_0 = {M0[0], M0[2], M0[4]};
LUT_L0_N0 LUT_L0_N0_inst (.M0(inpWire0_0), .M1(M1[0:0]));

wire [2:0] inpWire0_1 = {M0[1], M0[2], M0[3]};
LUT_L0_N1 LUT_L0_N1_inst (.M0(inpWire0_1), .M1(M1[1:1]));

wire [2:0] inpWire0_2 = {M0[0], M0[1], M0[2]};
LUT_L0_N2 LUT_L0_N2_inst (.M0(inpWire0_2), .M1(M1[2:2]));

endmodule
\end{lstlisting}

\begin{lstlisting}[language=Verilog, caption=LUTL0N0.v, label=LUTL0N0v]
module LUT_L0_N0 ( input [2:0] M0, output [0:0] M1 );
        reg [0:0] M1;
        always @ (M0) begin
                case (M0)
                        3'd0: M1 = 1'b1;
                        3'd1: M1 = 1'b1;
                        3'd2: M1 = 1'b1;
                        3'd3: M1 = 1'b0;
                        3'd4: M1 = 1'b1;
                        3'd5: M1 = 1'b0;
                        3'd6: M1 = 1'b0;
                        3'd7: M1 = 1'b0;
                endcase
        end
endmodule
\end{lstlisting}

\begin{lstlisting}[language=Verilog, caption=LUTL0N1.v, label=LUTL0N1v]
module LUT_L0_N1 ( input [2:0] M0, output [0:0] M1 );
        reg [0:0] M1;
        always @ (M0) begin
                case (M0)
                        3'd0: M1 = 1'b1;
                        3'd1: M1 = 1'b0;
                        3'd2: M1 = 1'b1;
                        3'd3: M1 = 1'b0;
                        3'd4: M1 = 1'b1;
                        3'd5: M1 = 1'b0;
                        3'd6: M1 = 1'b1;
                        3'd7: M1 = 1'b0;
                endcase
        end
endmodule
\end{lstlisting}

\begin{lstlisting}[language=Verilog, caption=LUTL0N2.v, label=LUTL0N2v]
module LUT_L0_N2 ( input [2:0] M0, output [0:0] M1 );
        reg [0:0] M1;
        always @ (M0) begin
                case (M0)
                        3'd0: M1 = 1'b1;
                        3'd1: M1 = 1'b0;
                        3'd2: M1 = 1'b1;
                        3'd3: M1 = 1'b0;
                        3'd4: M1 = 1'b1;
                        3'd5: M1 = 1'b0;
                        3'd6: M1 = 1'b1;
                        3'd7: M1 = 1'b0;
                endcase
        end
endmodule
\end{lstlisting}

\newpage

\begin{table}
    \begin{sidecaption}[Analytical vs.\ True LUT cost.]{%
        Comparing the Analytical LUT Cost formula with the results we actually get from synthesis of the network. Note that this is for a \textbf{purely combinational circuit} implementation of the Neural Network. 
    }[designautomate:tab:truelutcost]
\begin{threeparttable}
\begin{tabular}{lrr}
\hline
Analytical LUT cost & LUTs After Synthesis & Reduction     \\ \hline
128      & 80                & 1.6 $\times$  \\
272517   & 54336             & 5.01$\times$  \\
726180   & 76336             &  9.5$\times$  \\ \hline
\end{tabular}
\end{threeparttable}
\end{sidecaption}
\end{table}

\begin{table}
\begin{threeparttable}
\begin{tabular}{llllrrrrr}
\hline
\multicolumn{3}{c}{Model}      & Analytical LUTs & \multicolumn{5}{c}{Resources}    \\ \hline
X  & BW & HL                   &                 & LUT   & FF   & DSP & BRAM & WNS  \\ \hline
3  & 2  & 64, 32, 32           & 266             & 132   & 160  & 0   & 0    & 4.04 \\
5  & 2  & 64, 32, 32           & 5586            & 2132  & 393  & 0   & 0    & 2.54 \\
3  & 4  & 64, 32, 32           & 45220           & 7117  & 3394 & 0   & 5    & 1.073\\
7  & 2  & 64, 32, 32           & 90706           & 22146 & 1533 & 0   & 0    & 1.073\\
4  & 3  & 64, 32, 32, 32, 32   & 50235           & 16338 & 1329 & 0   & 5    & 1.44 \\ \hline
\end{tabular}
\end{threeparttable}
\begin{caption}[Model Resource Costs]{%
    Analyzing the result of synthesizing neural networks trained on the Jet Substructure Classification problem by FPGA4HEP, using different Neuron Fan-Ins. The synthesized networks have registers between layers and at the input. Note that the clock target was 5ns. 
}
\label{designautomate:tab:fpga4hepwithregs}
\end{caption}
\end{table}

\section{Logic Synthesis and Analytical LUT estimates}
Logic synthesis is a key stage in the computer aided design flow of a FPGA. It is composed of a series of optimizations to improve the performance of the design. To support logic synthesis, our tool-flow is fairly straight forward. Once the VERILOG Generator has generated all the codes for each neuron, we can simply import the $synthesize\_and\_get\_resource\_counts$ function from $fpganet.synthesis$. Once called, it returns a list of all the hardware resource that are needed. It also generates a log file which contains a verbose description of the synthesized model. 

Upon synthesizing a few neural networks on Vivado using the Logic Synthesis tool-flow, we observed that the Analytical LUT cost was an overestimation. This can be seen in \cref{designautomate:tab:truelutcost}. In general, we observed that our Logic Synthesis tool optimized the implementation such that we only need a fraction of the actual estimate. This raises very interesting question about the topologies we can explore. We had restricted ourselves in topology exploration due to certain fan-ins being beyond the scope of fitting on any FPGA fabric. \\
The networks we have reported in later sections are all using our Analytical LUT Cost model, which is actually an overestimation as is evidenced by \cref{designautomate:tab:truelutcost}. It seems that as the Analytical LUT cost increases, the reduction in the true LUT resource needed is more notable, thought this claim needs a lot more testing to be proven empirically. \\
This LUT costs presented in \cref{designautomate:tab:truelutcost} only present purely combinational implementation of the neural network. After observing the cost of such an implementation, we need to look at the how the resource cost behaves when we have registers at intermediate stages (between layers). We also need to conduct some timing analysis of the circuit generated by this process.

\subsection{Resource cost with and without Registers in circuit}
As we see in \cref{fig:fpganetmodule}, between every LUTLayer the activations are wired to intermediate registers, this is also true for the input of the LogicNetModule. In this section we previously looked at the LUT cost reduction after synthesis without any registers (a purely combinational circuit). The LUT cost reduction for a network without registers between intermediate layers could be greater, due to the fact that now the optimization problem does not have a hierarchy (across layers). This could also cause an increase in the synthesis time. We observed a noticeable decrease in synthesis time for a circuit with registers, but do not provide empirical evidence to back this observation.

\section{Timing Analysis}
In order to do some timing analysis on a basic LogicNet topology, we synthesized a small network by creating a fully-pipelined VERILOG description. After running synthesis, placement and Routing using Vivado, we observed a resource usage of 150 LUTs (from the analytical cost of 212 LUTs), and a minimum clock period of $0.768$~ns (frequency of 1.3~GHz), due to a very small circuit size. This indicates that there is potential for further reduction of resource footprint using logic minimization. Also note that the maximum clock frequency supported by the global clock network on the FPGA tested is 666 MHz, which leaves a large amount of slack for larger topologies. This actually stands true for more complicated topologies that we tested. These topologies are listed in \cref{designautomate:tab:fpga4hepwithregs}, the WNS stands for the 'Worst Negative Slack'. Slack is defined as the difference between the actual and the desired time for a timing path. A good FPGA design should have zero slack, if we have negative slack then the timing constraints we have specific for our design is not met. In such a scenario we can either relax our constraints or come up with a better design to meet these timing constraints we had initially specified. All the reported synthesis experiments in \cref{designautomate:tab:fpga4hepwithregs} are with a clock target of 5ns. For the first model in the table, this results in a frequency of $\frac{1000}{5-WNS} \approx 1042 MHz$. While it is evident that the frequency decreases as the LUT cost increases, we will also be conducting experiments with a more strict 1ns clock target in the future as well as more in-depth analysis of the LUT cost with and without registers.\\

It is also very interesting to note that in larger topologies, the logic synthesizer not only uses the LUTs, but also stores some neurons in the BRAMs. As we had stored truth tables themselves as primitives, it is really interesting to see what 'Hardware Building Blocks' are generated to store the Neurons. While this is interesting, as we make the clock target more strict, the synthesizer may choose to purely use LUTs. This is yet to be observed and more testing needs to be done to understand how the analytical LUTs differ from the true resource allocation of the synthesizer. \\

\clearpage

\clearpage
\section{Research Recess II – Synthesis}

The implications of \cref{designautomate:tab:truelutcost} to topology design is of great importance. We believe that there are two important questions we need to answer.
\subsection{Can we design a heuristic that aids LUT cost reduction for a neuron?}
In this library, we train neurons with static random connectivity. While this idea leads to sub-optimal accuracy, we are not too concerned with this as we can adopt any pruning technique and convert that into a mask-weight tensor and port that to a LogicNet model. An interesting question to ask is, is there a way to train each \textit{neuron}, such that during synthesis the LUT cost is dramatically lower than the analytical LUT cost? \\
This question is unanswered in this thesis. However, in this section, we hope to delve into this to some extent and define a way we can incorporate this into the training procedure. Our LogicNet library supports truth table calculation. This can be a slow process, but if a model only has to be deployed for inference; we can propose a training scheme that does take this into account. \\

PyEDA is a python library for Electronic Design Automation. Logic minimization is known to be a NP-Complete problem. It supports Truth Table Minimization. Instead of minimizing expressions, we can use our Truth Table Generator and integrate that with PyEDA to get the minimized truth table. It is important to note that this will significantly increase the train time of the neural network. Thus, we keep it beyond the scope of this thesis. We still however, nudge the idea of creating a cost function which takes the truth table minimization into account. Studying the hyper-parameter setting for such a cost function and its effect on the accuracy would be a very interesting topic. 

\subsection{Can we integrate congestion estimation in an FPGA with neural architecture search for non-layered topologies?}
There has been a lot of interest around congestion estimation for FPGAs. In a non-layered scenario, it could be really important to take congestion of an FPGA into account. The most efficient methods are based on fast-to-compute heuristics \cite{Swartz:1998:FRR:275107.275134} but they often have poor estimates. \cite{Yeager2007CongestionEA} uses a global router during placement, which gives more accurate congestion estimates, but at the cost of runtime. \\
The aforementioned methods provide estimates during place-and-route. Above a certain complexity of topology design, we may need to develop a heuristic which allows us to give a congestion estimate for a certain topology. This is a very complicated research problem, but there has been some progress in this direction. \cite{Samajdar2019ScalingTC} design directly in RTL component-level instantiations of DSP, BRAMs and URAMs and associated controller's for orchestrating data movement and manually leverage dedicated cascade interconnects for high frequency, nearest-neighbour data movement. \\
Given a netlist representation of a circuit, the elementary objective of congestion minimization is to minimize the total sum of wirelength for each net. Perhaps, it may also be a better endeavor to integrate certain design principles into topology constraints instead of attempting to get a real congestion estimation. This remains in our opinion an interesting research problem. It is a hard problem to formalize, and may be very interesting to implement. 
\chapter{LogicNet4HEP}\label{ch:fpganet4hep}
\openepigraph{The Europeans and the Americans are not throwing \$10 billion down this gigantic tube for nothing. We're exploring the very forefront of physics and cosmology with the Large Hadron Collider because we want to have a window on creation, we want to recreate a tiny piece of Genesis to unlock some of the greatest secrets of the universe.}{Michio Kaku}
The LHC (Large Hadron Collider) is the worlds highest energy particle accelerator. Proton bunches collide at a frequency of 40 MHz, with data-rates in excess of hundreds of terabytes per second. The process of tagging and filtering data at real time is called triggering. Triggering helps reduce data to manageable levels for processing offline. At the CMS detector, this triggers is performed at two stages. The first stage typically uses custom hardware like ASICs or FPGAs~\cite{Duarte_2018} to handle the initial data rates with latency in the range of a few hundres of nano seconds. The second level of triggering, known as High Level Triggering (HLT) uses commercial CPUs to process filtered data in software with longer processing times. We borrow some definitions from the work of Javier Duarte\cite{Duarte_2018} to formalize the task in this section.\\

\marginpar{\centering
\includegraphics[width=70pt]{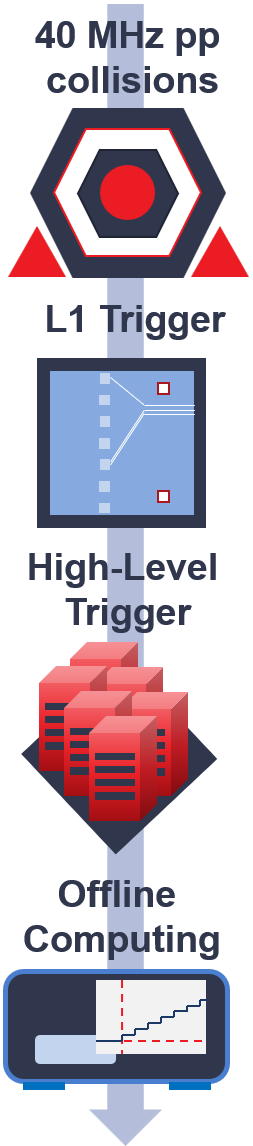}
\captionof{figure}{The LHC big data problem.}
\label{fig:lhcdataflow}
}

Jets are collimated showers of particles that are created from the decay and hadronziation of quarks $q$ and gluons $g$. Due to the high collision energy at LHC, jet signatures of interest emergy from overlapping quark-initiated showers produced in the decays of heavy standard model particles.
In this task, we focus on the task of classifying a jet as either a quark ($q$), gluon ($g$), W boson ($W$), Z boson ($Z$) or top quark ($t$) jet. 
\newthoughtpar{Key Metrics}
To create an optimal L1 trigger that satisfies the require  ments of the target algorithm, we list the three key metrics considered by~\cite{Duarte_2018}. 
\begin{itemize}
    \item \textbf{Latency}: The total time (usually in units of "clocks") for a single forward pass through the neural network.
    \item \textbf{Initiation Interval}: The number of clock cycles before the neural network can accept a new input. Inference rate is inversely proportional to the initiation interval. This is of key importance as the data should be pipe-lined at the rate of the initiation interval.
    \item \textbf{Resource Usage}: This may encompass the counts of a range of components available on the target FPGA fabric. Some of which may be Block RAMs (BRAMs), Digital Signal Processing Blocks (DSPs), Registers, Lookup-Tables (LUTs).
\end{itemize}
\newthoughtpar{Models}
As LogicNet style topologies are quite unique in the heavy sparsity and quantization they desire, we had to do extensive topology exploration to gain insights into how we can get the most performant networks with the lowest resource cost.
\newthoughtpar{Neural Architecture Search}
We considered Neural Architecture Search to find efficient topologies. Neural Architecture Search is a method for automating the design of Neural Networks. There are many ways to define the search space, strategy and getting a performance estimation for evaluating a discovered architecture. In our case, it makes sense to treat neurons as individiual blocks, and discovering not only the hierarchical connectivity, but aiming to discover completely unstructured (not layered) topologies. On top of this, we also have to define a search space carefully. Here, the search space is not only limited to the bit-width and connectivity per neuron but also the overall connectivity across layers. The Performance estimation would fairly straightforward, as we have estimates of the resource usage, as well as can easily give an account of the initiation interval and latency. The initiation interval for an LogicNet style topology is 1, and the latency is essentially equal to the number of intermediate activations that are generated, in a layered fashion.
However, due to the combinatorial nature of this problem, we were too constrained to do a Neural Architecture Search. We aimed to prove that LogicNet style structures can deliver performance benefits. It is very much within the scope of future research to integrate NAS into the LogicNet library, and the library itself is being built with the same in mind. 
\begin{figure}[h]
    \centering
    \includegraphics[width=300pt]{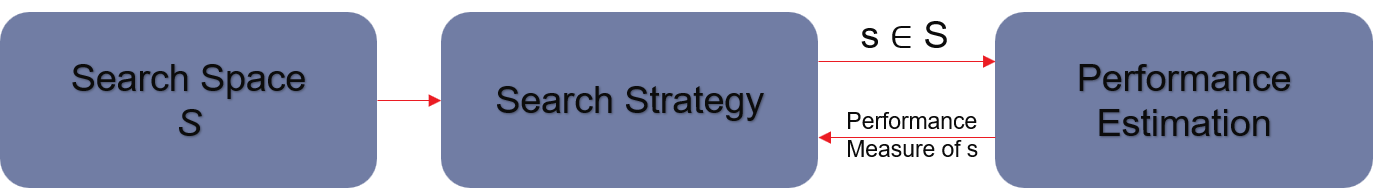}
    \caption{Dimensions of the NAS method.}
    \label{fig:NAS}
\end{figure}

\newthoughtpar{Axes of exploration}
As we are exploring LogicNet type topologies by hand, it becomes very important to specify the 'axes of exploration', and make observations along the same lines. For us, there were 3 primary axes of exploration. The Bit-Width of activation, the number of neurons per hidden layer, and the Fan-In (In Bits) per neuron. We have not attempted to explore variable fan-in for neurons with-in a layer. We do however give insights into variable fan-in for individiual layers.

\marginpar{\centering
\includegraphics[width=90pt]{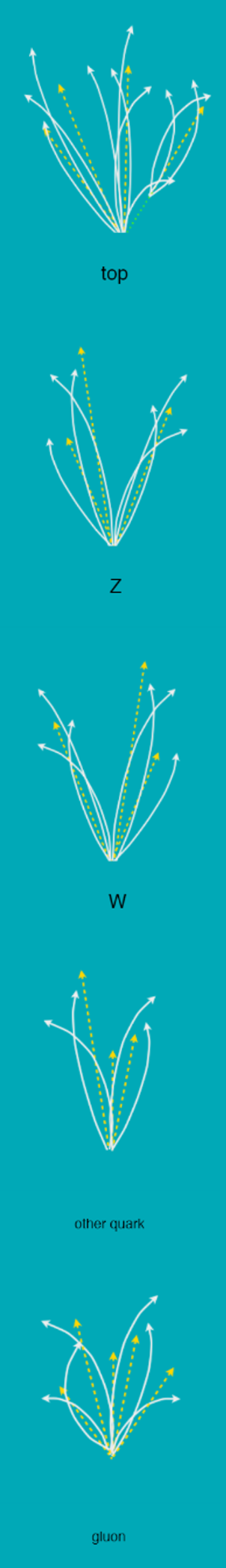}
\captionof{figure}{Discriminating between highly energetic (boosted) \textbf{\textit{q, g, W, Z, t}} initiated jets. Figure adapted from User: FPGA4HEP on GitHub.}
\label{fig:classesFPGA4HEP}
}

\newpage

\begin{figure}[h]
    \centering
    \includegraphics[width=300pt]{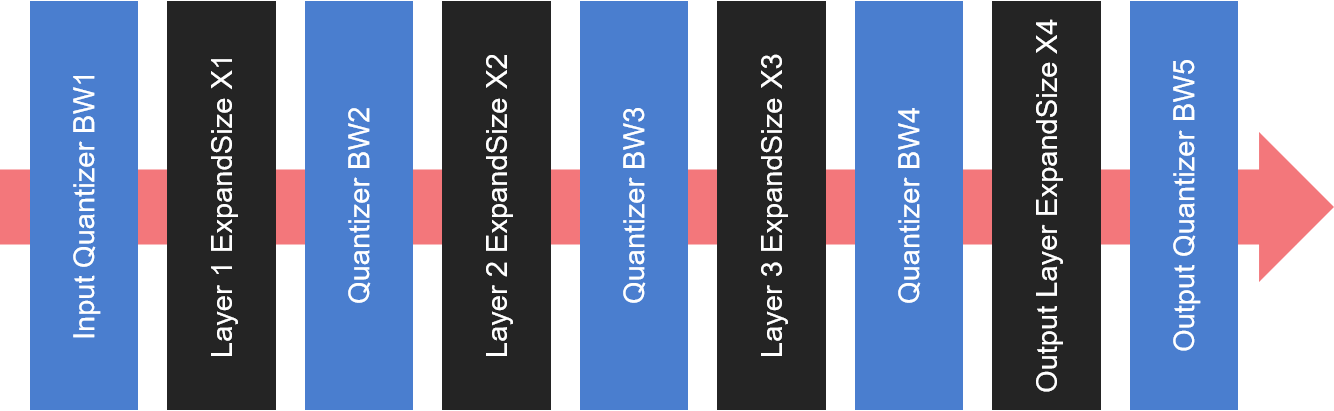}
    \caption{Our PFGA4HEP Neural Network Architecture.}
    \label{fig:FPGA4HEPtopology}
\end{figure}

\newthoughtpar{Classification performance}
To quantify the performance of the classifier we have made in ~\cref{fig:FPGA4HEPtopology}, we use the same metric sued in~\cite{Duarte_2018}. We use the AUC metric, or area under the Receiver Operating Characteristic (ROC) curve. The ROC curve is given by the background rejection versus signal efficiency computed from sequential cuts on the classifier output, where background rejection is (1 - background efficiency). The Expected AUC is the AUC acheived by a a full 32-bit floating point inference of the neural network. 

\begin{figure}[h]
    \centering
    \includegraphics[width=320pt]{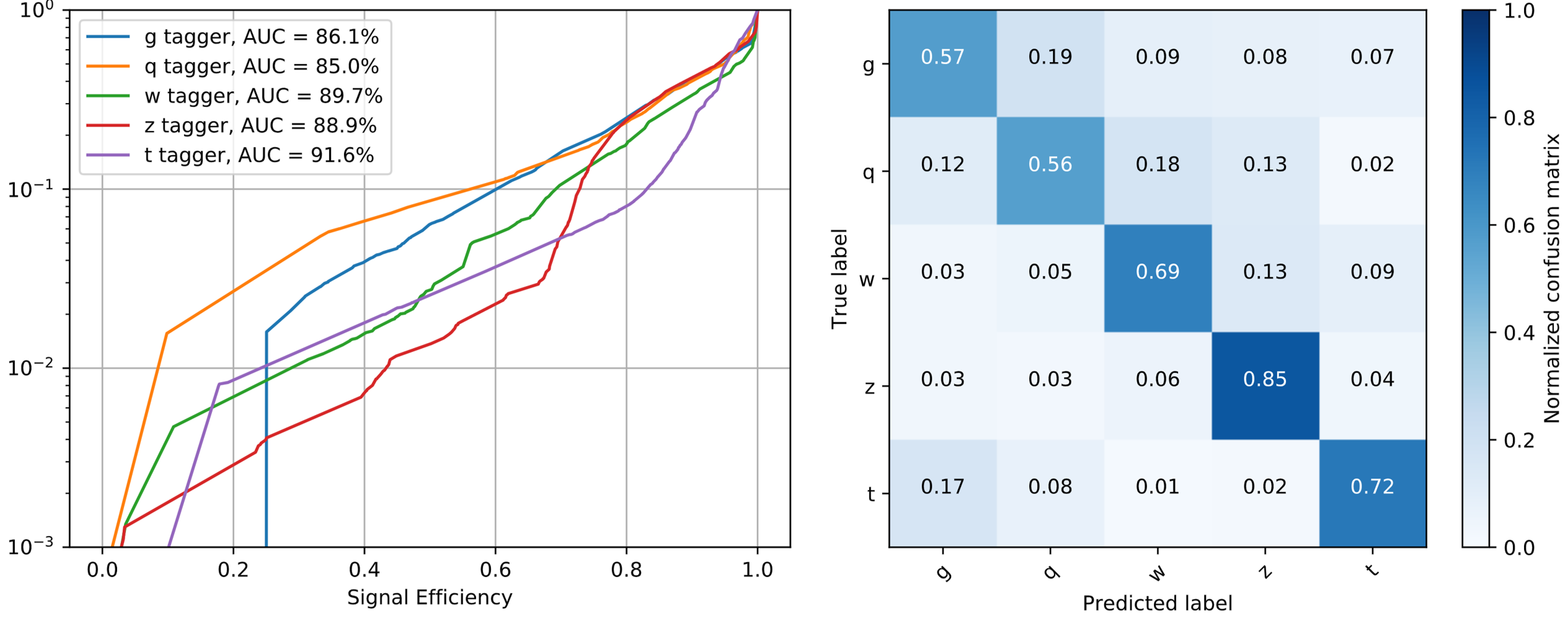}
    \caption{Performance of the Deep Neural Network Classifier. The Left figure shows the signal efficiency versus the mis-identification rate for q, W, Z, t, g jet identification. The mis-identification rate is based on an equal admixture of the other non-signal jet types. On the right we show the Normalized Confusion Matrix for the classifier.}
    \label{fig:classifierperform}
\end{figure}

\begin{figure}[h]
    \centering
    \includegraphics[width=400pt]{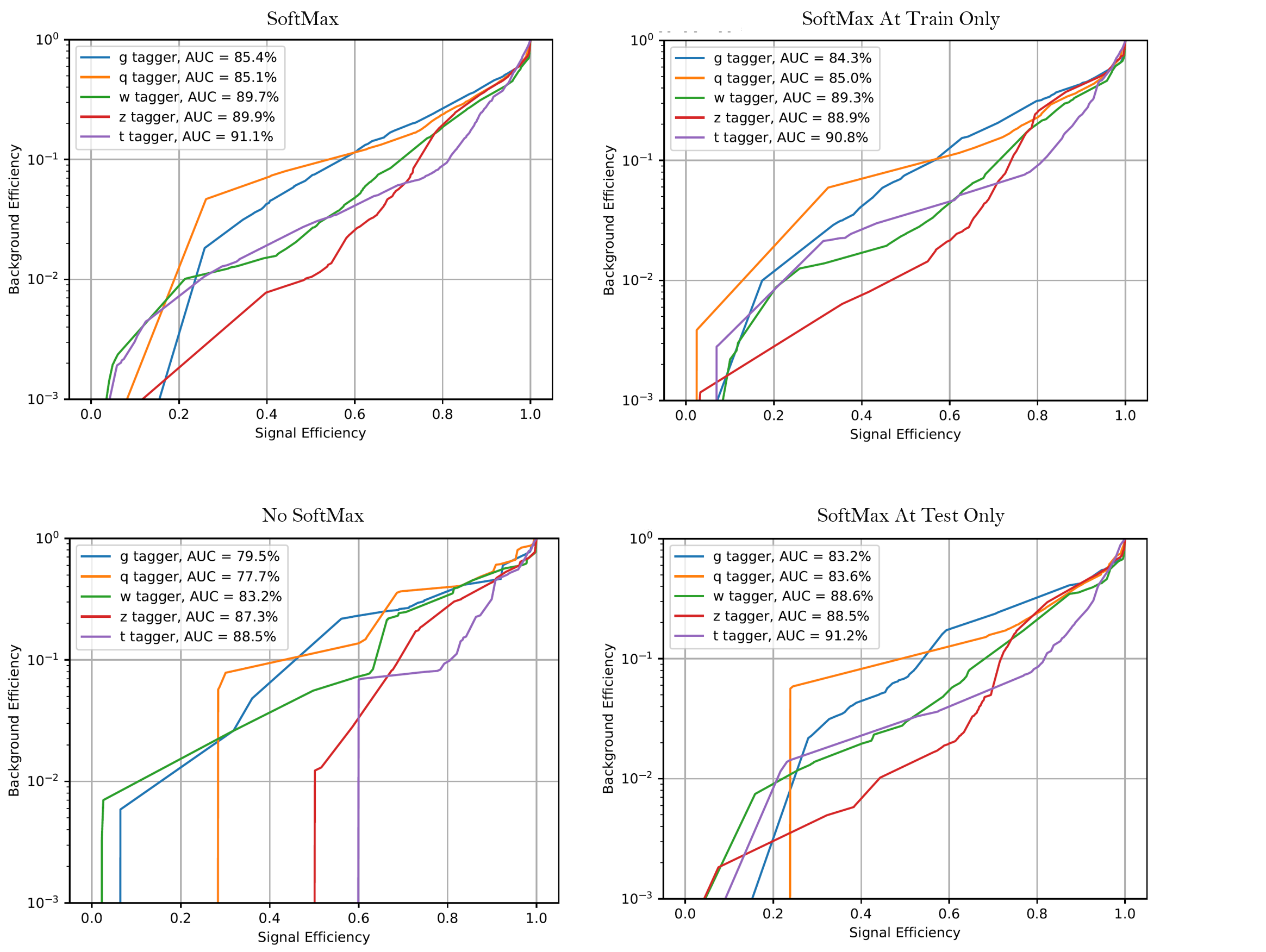}
    \caption{We train a model on the FPGA4HEP dataset and observe the classifiers Receiver Operating Characteristics with respect to the SoftMax Layer.}
    \label{fig:softmaxtests}
\end{figure}

\newthoughtpar{Model Descriptions}
It is important to know how we are labelling our models. We refer the users to \cref{fig:FPGA4HEPtopology}. Our models are described in \cref{fpganet4hep:tab:fpga4hepmodels}. The $2^{nd}$ (HL) column can be referred to as $(Neurons_{Layer1}, Neurons_{Layer2}, Neurons_{Layer3})$. The BW refers to the Bit-Width, with X indicating the fan-in of neurons. Note that this fan in is not in bits, but instead the number of 'synapses'. The last layer is special, its bit-width and fan-in are very crucial. Thus, we have specific the fan-in of some models with $X_{fc}$. Where this was not been specified, the final layer would be dense. $BW_{fc}$ refers to the bit-width of the output. This is important as lower bit-width and removal of the SoftMax layer generally results in degradation of the ROC curve. However, the extra cost of a SoftMax layer has been excluded from our analysis.



\begin{table}
    \begin{sidecaption}[FPGA4HEP Model List]{%
        Model descriptions
    }[fpganet4hep:tab:fpga4hepmodels]
\begin{threeparttable}
\begin{tabular}{lrrrrrrrrr}
\hline
Model & HL            & BW & X & $X_{fc}$ & $BW_{fc}$   & LUTL1 & LUTL2 & LUTL3 & LUTL4 \\ \hline
A     & (64, 64, 64)  & 3  & 3 & -         & 3          & 2112  & 2112  & 2112  & 4125  \\
B     & (128, 64, 32) & 3  & 3 & -         & 3          & 4224  & 2112  & 1056  & 2090  \\
C     & (64, 32, 32)  & 2  & 3 & -         & 2          & 128   & 64    & 64    & 1415  \\ 
D     & (64, 32, 32)  & 2  & 5 & 6         & 4          & 2688  & 1344  & 1344  & 3400  \\ 
E     & (64, 64, 64)  & 2  & 4 & 4         & 4          & 640   & 640   & 640   & 200   \\\hline 
\end{tabular}
\end{threeparttable}
\end{sidecaption}
\end{table}

\begin{table}
    \begin{sidecaption}[FPGA4HEP Main Table]{%
        Accuracy and LUT costs of some explored models. Note that the LUTs is the estimate described by \eqref{lutcostcloseform}. An in-depth discussion of the closed form equation of cost versus the synthesized cost has been discussed before. Note that the \textbf{g, q, W, Z, t} are reported as Area Under Curve - Receiver Operating Characteristic.
    }[fpganet4hep:tab:fpga4hepresults]
\begin{threeparttable}
\begin{tabular}{lrrrrrrrrrrrrrrrr}
\hline
Model &    g &   q  &   W  &   Z  &   t  & Avg AUC-ROC  & LUTs     & \% FC \\ \hline
A     & 89.3 & 86.2 & 89.8 & 89.3 & 92.7 & 89.46  & 10461    & 39.43 \\
B     & 86.5 & 86.0 & 90.5 & 90.5 & 91.9 & 89.08  & 9482     & 22.04 \\
C     & 85.4 & 82.9 & 84.9 & 83.1 & 90.3 & 85.32  & 1671     & 84.68 \\ 
D     & 85.8 & 85.0 & 89.5 & 89.2 & 91.5 & 88.20  & 8776     & 38.74 \\ 
E     & 86.5 & 85.4 & 90.0 & 89.6 & 92.0 & 88.70  & 2120     & 9.43  \\\hline 
    \end{tabular}
\end{threeparttable}
\end{sidecaption}
\end{table}


\begin{figure}[h]
    \centering
    \includegraphics[width=350pt]{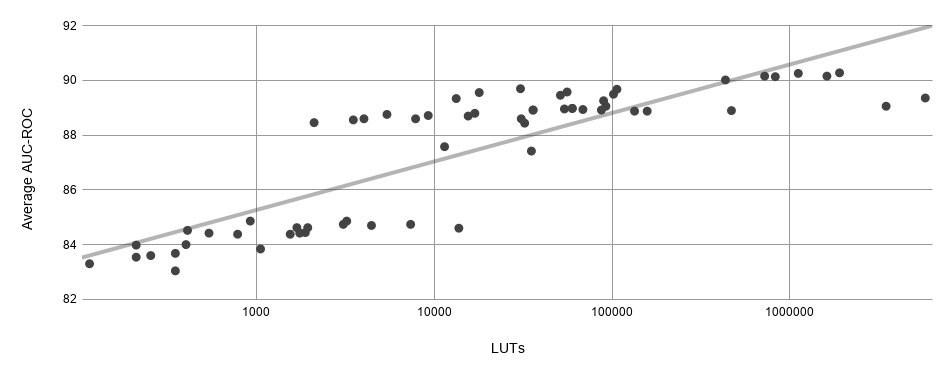}
    \caption{Increase in Accuracy with increase in LUT cost.}
    \label{fig:lutvacc}
\end{figure}

\begin{figure}[h]
    \centering
    \includegraphics[width=350pt]{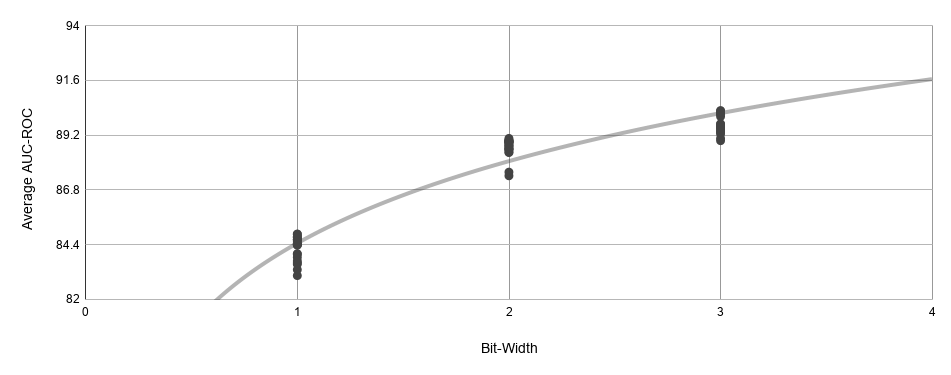}
    \caption{Increase in Accuracy with increase in Bit-Width.}
    \label{fig:BWvacc}
\end{figure}

\newthoughtpar{Performance with and without SoftMax}
In the previous section, we described our models and briefly touched upon the effect of SoftMax on ROC. Not only does the SoftMax make the training better, it makes the classifier more robust to noise. While the effects of removing the SoftMax layer after training is not evident on the confusion matrix, upon graphing the ROC, it becomes important to inspect it more closely.

\newthoughtpar{ROC - Curve and SoftMax}
To test this more closely, we train a model in two ways. The first test is to train a neural network with output quantization, but without SoftMax. We then find its ROC and compare it with the ROC of the same trained model with a SoftMax appended after the final quantizer.The second test is to train a neural network with output quantization followed by a SoftMax. We will then find its ROC and compare it with the ROC of this same trained model but we will drop the final SoftMax layer. This will lend us insights as to whether there is any train time benefits of the final SoftMax layer, additionally also allow us to make claims on whether removing the SoftMax layer from a trained model causes detriments in the ROC of the classifier. 

\newthoughtpar{Exploring the Axes}
To gain further insights into the effects of architectural decisions on the accuracy and LUT costs, we did some elementary grid search.
In Figure \cref{fig:lutvacc}, it is fairly evident that increasing the LUT cost gives higher accuracies. We do however see a clear overlap in two distinct clusters of 'Average AUC-ROC', whereby some models that need very little luts (as less as 2500 LUTs) perform as well as models requiring about 100000 LUTs. Further, we observe that there are some models that require a million LUT, but barely surpass the performance of a model need as little LUTs as 2500. Upon inspection, we discovered that these were models with unreasonably large final layers, or topologies where the per neuron fan-in in bits is too large, while focusing very little on the relationship between Bit-Width and Expand-Size for accuracy.
This becomes more evident in Figure \cref{fig:BWvacc}, where we see that while going from a Bit-Width of 1 to 2 almost undoubtedly reaps benefit. We see diminishing returns when going from bit-width of 2 to 3. While it is not conclusive that increasing bit-width consistently results in diminishing returns, it is still an important insight that we must be aware of the relation between Bit-Width and ExpandSizes. An unwise compromise on one may give sub-optimal results for the same number of LUTs. 

\newthoughtpar{Pruning Techniques}
As discussed in a previous chapter, we have explored many different pruning techniques. While iterative pruning gave the best accuracies, A-Priori Fixed Sparsity was the easiest to train. It thus becomes important to question what utility an Iterative Pruning strategy has over A-Priori Fixed Sparsity, and whether it has any notable benefits for topology exploration. It may be within reason to say that it is worth doing topology exploration using A-Priori Fixed Sparsity, and then taking the discovered topology and training it with Iterative Pruning. It is important to note that at this point of time, the LogicNet Library only supports A-Priori Fixed Sparsity. 
As we can see in \cref{fpganet4hep:tab:iterativeVpriori}, there is a very marginal difference in the \textbf{Average Area Under Curve of the Receiving Operating Characteristic}. 
It thus may be to our benefit to leave topology exploration to A-Priori Fixed Random Sparsity, and use more advanced pruning technique once a suitable topology has been discovered. This claim however is up for further introspection.

\begin{table}
    \begin{sidecaption}[Iterative vs A-priori Fixed Sparsity]{%
        This table depicts the accuracies for particular models trained with A-Priori Fixed Sparsity and Iterative Pruning techniques. 
    }[fpganet4hep:tab:iterativeVpriori]
\begin{threeparttable}
\begin{tabular}{lrrr}
\hline
Model & LUTs  & A-Priori Fixed Sparsity & Iterative Pruning \\ \hline
A     & 2120  & 88.46                   & \textbf{88.7}     \\
B     & 54060 & 88.96                   & \textbf{89.56}    \\
C     & 59840 & \textbf{88.98}          & 88.92          \\    \hline 
    \end{tabular}
\end{threeparttable}
\end{sidecaption}
\end{table}



\chapter{MNIST}\label{ch:mnist}

\openepigraph{AI is a tool. The choice about how it gets deployed is ours.}{Oren Etzioni}

The MNIST database of handwritten digits, available from this page, has a training set of 60,000 examples, and a test set of 10,000 examples. This is a well known data-set, and is targeted by both MLPs and Convolutional Neural Network. We thus found  this to be an excellent starting point for LogicNet. 
We do some topology exploration on this data-set, and study a wide array of architectural decisions in detail. \\

While exploring our search space, we begun with MLPs on the MNIST data-set. To gain insights, we had to ascertain a few things which would make our search more effective. Some of the notable observations while attempting to gain insights on how to discover good topologies were as follows. Note that the observations below are specific to the MNIST data-set and only valid for MLPs.
\begin{itemize}
    \item For MLPs, going from 2 hidden layers to 3 hidden layers still has benefits, i.e. the intermediate layers do not learn an identity function.
    \item In our tests, a drop-out at the input layer gives very slight benefits.
    \item Drop-out in the intermediate layers with A Priori Random Fixed Sparsity causes accuracy to deteriorate irrespective of the size of the hidden layer, or bit-width.
    \item ADAM and SGD as optimizers give very similar accuracy results.
    \item Increasing bit-width from 1 to 2 is beneficial, but not at the cost of neuron fan-in in terms of synapses. 
    \item Increasing number of neurons in a hidden layer is beneficial up-to 1024, after which the utility diminishes. 
    \item The last layer cannot have low per-neuron fan in. It causes serious detriments to accuracy of the network. For this reason unless mentioned otherwise; \textbf{all our experiments for MNIST report models with the final layer dense}.
\end{itemize}

\begin{figure}[h]
    \centering
    \includegraphics[width=330pt]{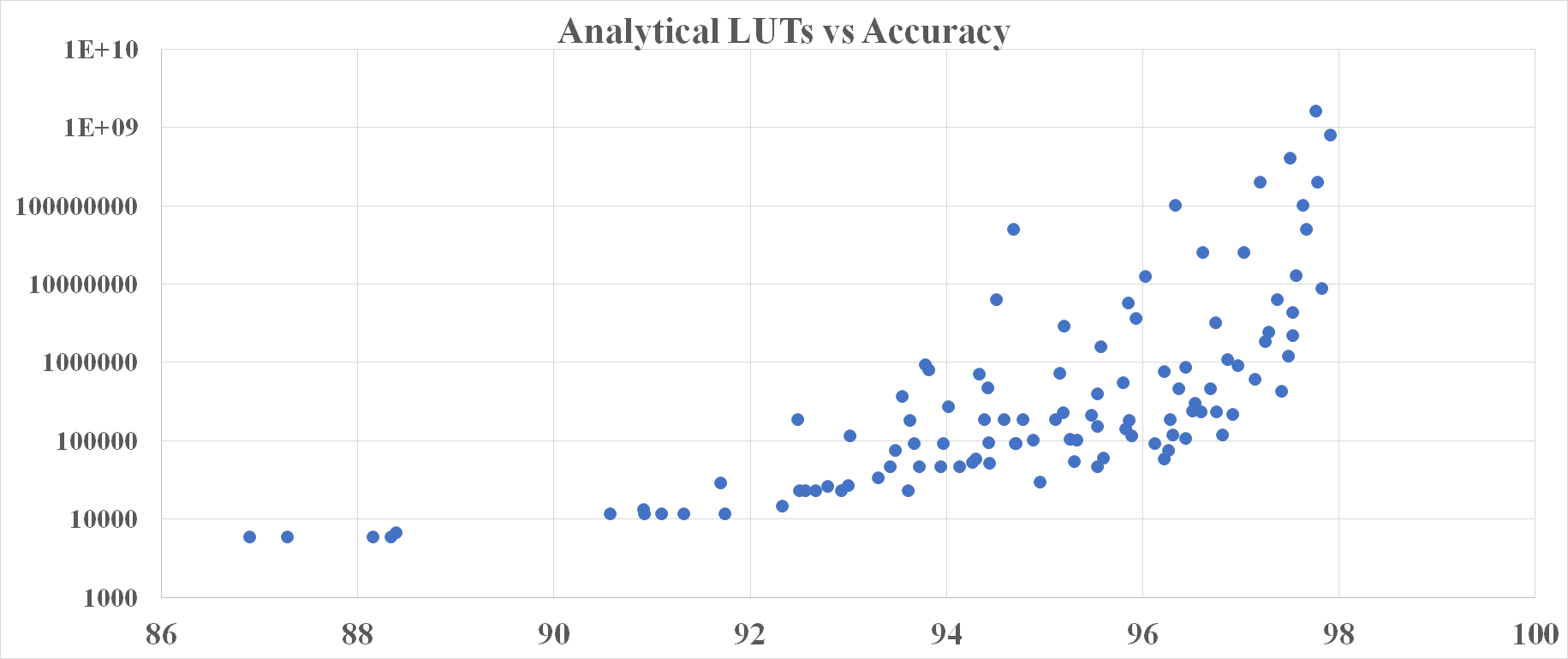}
    \caption{Analytical LUT cost vs. Accuracy for a 3 Layer MLP on the MNIST Data-Set.}
    \label{fig:3lmlplutacc}
\end{figure}

To study how our Analytical LUT cost varies with accuracy, we trained an array of models on the MNIST dataset, all of which were 3 Layer MLPs. \cref{fig:3lmlplutacc} Shows that there is consistently a lower bound in terms of LUTs for the same performance. It is worth nothing that the Y axis is logarithmic, which means that a poor choice of Neuron Fan-In with a sub-optimal Bit-Width and Connectivity configuration can be very detrimential. 

\begin{table}
    \begin{sidecaption}[MNIST MLP Model List]{%
        This table describes the models in detail, it also gives the layerwise breakdown of LUTs (From the Analytical LUT Cost Model) and the accuracy. These models have been trained with Fixed A-Priori Sparsity, and is just for topology exploration. 
    }[mnist:tab:mnistmodels]
\begin{threeparttable}
\begin{tabular}{lrrrrrrrr}
\hline
HL              & BW & X & LUTL1   & LUTL2 & LUTL3 & LUTL4    & LUTs & Accuracy \\ \hline
(512)$\times$1  & 2  & 6 & 87k   & 43.9k & -  & -           & 131k   & 95.62    \\
(1024)$\times$1 & 2  & 5 & 43k   & 87.7k & -  & -           & 130k   & 95.86         \\
(2048)$\times$2 & 2  & 5 & 86k   & 175k    & -    & -     & 261k     & 94.77          \\ 
(512)$\times$2  & 2  & 6 & 87k   & 87k    & 43.5k    & -    & 217.6k & 96.58     \\ 
(1024)$\times$2 & 2  & 5 & 43k   & 43k    & 87k    & -     & 173k    & 96.85         \\ 
(2048)$\times$2 & 2  & 5 & 86k   & 86k    & 174k    & -     & 345.8k & 96.96         \\ 
(512)$\times$3  & 2  & 6 & 87k   & 87k    & 87k    & 43.5k  & 304.6k & 96.53   \\ 
(1024)$\times$3 & 2  & 5 & 43k   & 43k    & 43k    & 87k  &  215.9k  & 96.92    \\ 
(2048)$\times$3 & 2  & 5 & 86k   & 86k    & 86k & 173.8k   &  431.8k & 97.41          \\ \hline
\end{tabular}
\end{threeparttable}
\end{sidecaption}
\end{table}

\begin{figure}[h]
    \centering
    \includegraphics[width=330pt]{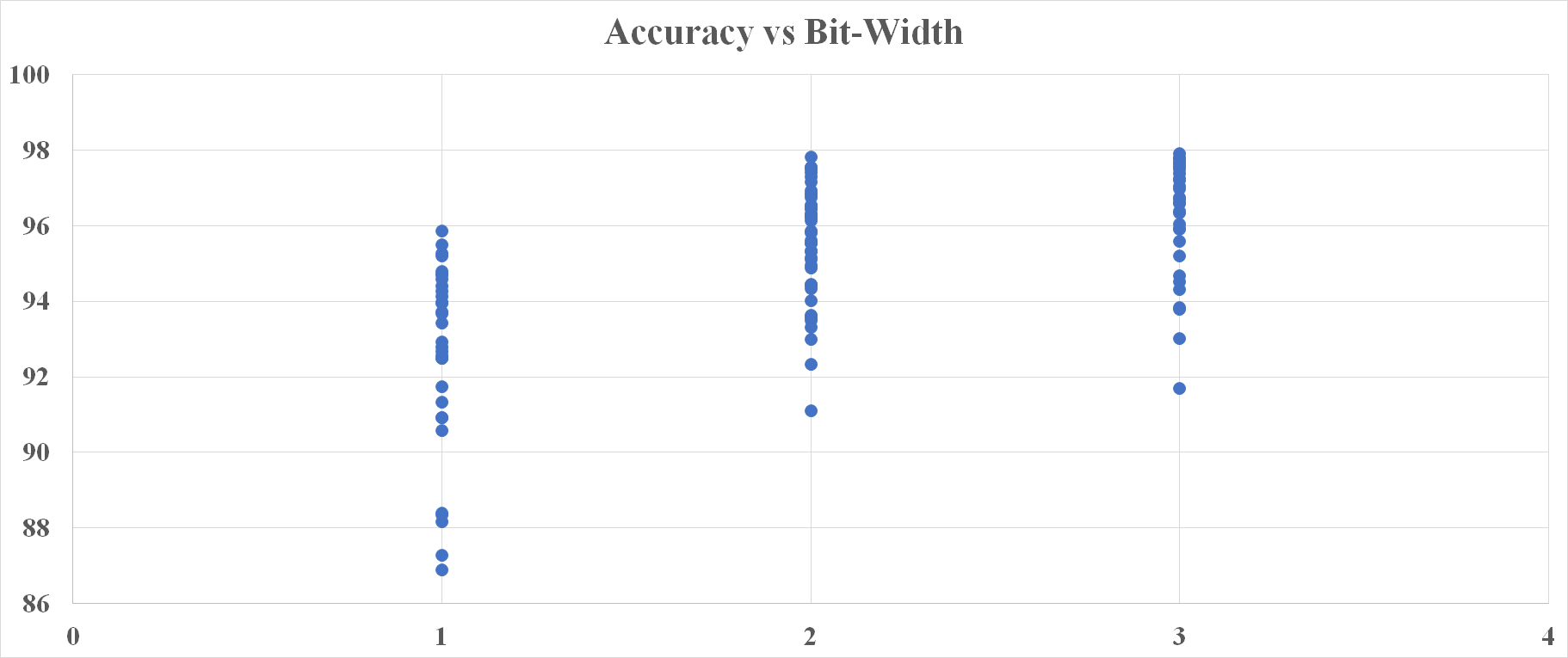}
    \caption{Accuracy vs. Bit-Width for a 3 Layer MLP on the MNIST Data-Set.}
    \label{fig:3lmnistaccbw}
\end{figure}

From \cref{mnist:tab:mnistmodels}, we gain further insights into the role of depth and width of the network for accuracy on MNIST. We see an upward trend in accuracy as we increase the depth of the neural network. Further, we can see the benefits of increasing the Bit-Width for a 3 Layer MLP on the  MNIST data-set from \cref{fig:3lmnistaccbw}. \\


\newthoughtpar{Iterative and Learned Sparsity}
In our tests, A-Priori Fixed Random Sparsity consistently gave poor accuracies when compared to Iterative and Learned Sparsity. This may be due to the fact that the MNIST data-set is flattening an Image and thus has definite structure in the flattened input. Initializing random connectivity is thus detrimential, whereas iterative and learned sparsity methods are able to discover important connections. From \cref{mnist:tab:iterativevmomentumvrandom}, we can be relatively confident that Iterative Pruning is the best strategy to train models with such sparsity. Momentum Sparsity gives better results than A-Priori Fixed Sparsity and requires almost the same amount of resources and time. Iterative Pruning takes about 10$\times$ longer to train, based on your pruning rates.

\begin{table}
    \begin{sidecaption}[Accuracy vs Pruning Techniques]{%
        This table depicts the accuracies for models trained with A-Priori Fixed Sparsity, Momentum Sparsity and Iterative Pruning techniques. 
    }[mnist:tab:iterativevmomentumvrandom]
\begin{threeparttable}
\begin{tabular}{lrrrr}
\hline
Model & A-Priori Fixed Sparsity & Momentum Sparsity & Iterative Pruning \\ \hline
A     &  97.32             &          97.68        & \textbf{97.78}     \\
B     &  97.12             &          97.39    & \textbf{97.75}    \\
C     &  96.58             &          97.22    & \textbf{97.63}          \\    \hline 
    \end{tabular}
\end{threeparttable}
\end{sidecaption}
\end{table}

\newthoughtpar{Skip Connections}
Due to the nature of LogicNet, we can introduce 'Skip Connections' to a neural network. As long as the per neuron fan-in remains the same, the LUT cost remains the same. What influence this has on Place and Route and the complexity it introduces is beyond the scope of this thesis. \\
As evidenced by \cref{mnist:tab:skiparch}, we see notable increase in accuracy with  no overhead in LUT cost. Thus, this can be an important avenue to explore in the future.

\begin{table}
    \begin{sidecaption}[Skip Connections on MLPs]{%
        We observe the accuracy of 3 Layer MLPs with Skip Connections. 
    }[mnist:tab:skiparch]
\begin{threeparttable}
\begin{tabular}{lrrrr}
\hline
Model & No Skip & 1 Skip & 2 Skips            \\ \hline
A     &  92.86  & 93.89  & \textbf{94.66}     \\
B     &  94.47  & 95.03  & \textbf{95.18}     \\
C     &  95.71  & \textbf{95.88}  & 95.78      \\
D     &  94.67  & 95.24  & \textbf{95.51}      \\    \hline 
    \end{tabular}
\end{threeparttable}
\end{sidecaption}
\end{table}

\newthoughtpar{Convolutional Neural Networks}
To get better performance, we need to use topologies which are tailored for images. An exploration of how such sparsity translated to computer vision tasks is very important. As discussed before, we use 'Sparse Depthwise Separable Convolution' to facilitate realistic mapping of convolutions to LUTs. We also use heavy quantization. \cref{mnist:tab:degrademnist} Shows how accuracy is affected as we change the topology and its bit-precision and sparsity. We test the accuracy of 4 variants of the same topology. The 'FP' being the full precision model with vanilla convolutions, 'FP\_DW' being the full precision model but with Depthwise Separable Convolutions. 'FP\_X\_DW' further introduces sparsity in the depthwise separable convolution and finally 'QUANT\_X\_DW' quantizes the sparse depthwise separable convolutional model. \cref{mnist:tab:degrademnist} seems to indicate that quantization causes the most notable degradation in accuracy. 

\begin{table}
    \begin{sidecaption}[Bringing Convolutions to LogicNet]{%
        Taking a specific model and observing its accuracy as we change the Pruning, Quantization and Convolution method.
    }[mnist:tab:degrademnist]
\begin{threeparttable}
\begin{tabular}{lrrrrrrrr}
\hline
Model        & A     & B     & C     \\ \hline
FP           & 98.68 & 98.96 & 99.15 \\
FP\_DW       & 98.34 & 97.7  & 98.6  \\
FP\_X\_DW    & 97.25 & 97.93 & 97.41 \\
QUANT\_X\_DW & 96.84 & 97.59 & 95.89 \\ \hline
\end{tabular}
\end{threeparttable}
\end{sidecaption}
\end{table}

\begin{table}
    \begin{sidecaption}[Accuracy and LUT cost of CNNs on the MNIST data-set]{%
        This table describes the models in detail, it also gives the Analytical LUT Cost and the accuracy. These models have been trained with Fixed A-Priori Sparsity, and is just for topology exploration. BW Stands for Bit-Width, X stands for the (Xk, Xs). Xk is the kernel sparsity, and Xs is the sparsity of the pointwise operation in Depthwise Separable Convolution.
    }[mnist:tab:convmnist]
\begin{threeparttable}
\begin{tabular}{lrrrr}
\hline
Model & BW & X      & LUTs & Accuracy \\ \hline
A     & 2  & (5,5)  & 145k & 95.87    \\
B     & 2  & (3,5)  & 345k & 97.47    \\
C     & 2  & (5, 4) & 769k & 97.59    \\
D     & 2  & (5, 6) & 420k & 97.57    \\ \hline
\end{tabular}
\end{threeparttable}
\end{sidecaption}
\end{table}


\newthoughtpar{Convolution Skip Connections}
After testing skip connections on MLPs, it only made sense to explore what its scope is for convolutional neural networks. We present our findings similar to before, with 1 skip connection and 2 skip connections. The activation of the first convolutional layer and second convolutional layer is concatenated to the activation of the second convolutional layer and activation of the final convolutional layer respectively. 
From \cref{mnist:tab:convskip}, we see some accuracy improvements. We do not discuss these skip connections in too much detail as it is beyond the scope of this thesis. It is however a direction worth exploring. 

\begin{table}[h]
    \begin{sidecaption}[Skip Connections on MNIST Convolutional models.]{%
        We observe the accuracy of a 3 Layer LogicNet style Convolutional Neural Network with Skip Connections. 
    }[mnist:tab:convskip]
\begin{threeparttable}
\begin{tabular}{lrrrr}
\hline
Model & No Skip & 1 Skip & 2 Skips            \\ \hline
A     & 96.74 & 96.93  & \textbf{96.94 }    \\
B     & 97.08 & 97.56  & \textbf{97.57}     \\
C     & 96.94 & \textbf{97.36}  & 97.13      \\  \hline 
    \end{tabular}
\end{threeparttable}
\end{sidecaption}
\end{table}

\begin{fullwidth}
\part{Concluding Remarks}
\end{fullwidth}
\chapter{Conclusion}\label{ch:conclusion}
\openepigraph{The desire to create is one of the deepest yearnings of the human soul.}{Dieter F. Uchtdorf}

The neurons in our brains do not \textit{seem} to follow any given structure, rather they can be condensed to specialized graphs which are a far-cry from densely connected topologies we commonly use. We attempt to train extremely sparse structures, that can map directly to hardware building blocks in our FPGA. 
Much like we have taken jabs at trying to draw parallels between the neurons in our brains and the neurons in our machine learning framework, we take a jab at how to map neurons in these machine learning frameworks to hardware building blocks, we studied the interplay between quantization and sparsity. We developed a library to map hardware building blocks to neuron equivalents as well as used that library to synthesize neural networks. \\
We begin by introducing basic concepts such as the rough architecture of an FPGA, and some of the basic layer types in a Neural Network. We also delve lightly into Quantized Neural Networks and Sparse Neural Networks. In Chapter 2, we hope to explain how neural networks are typically implemented on FPGAs. We use the HLS-RFNoC workflow as an implementation case study. We then delve into the Analytical LUT cost that we use as our pessimistic resource cost for the rest of the paper.\\
Chapter 3 focuses on the exact process of mapping neurons to aforementioned 'hardware building blocks'. We explore different ways of sparsification of neural networks and delve into interesting research questions on implementing sparsity in neural networks. Here, we learn that A-Priori Random Fixed sparsity is an effective method for exploring topologies but we need more powerful methods such as Sparse Momentum Learning or Iterative Pruning to get the best accuracy. We also draw parallels between Sparsification and Neural Architecture Search, which is an interesting way to unify different tangents of research on discovering performant neural wiring.
We then move on to actually introducing to the reader the LogicNet design flow and particulars of the library itself. We explain the reasoning behind certain design decisions. Design automation is also discussed to some level, but in Chapter 5 we primarily focus on the components that are needed for design automation. We familiarize the reader with the truth tables we generate from a specific topology and delve in detail into how we implement our VERILOG code generator. We describe the modules and sub-structures that exist in the VERILOG code. This is followed by a very interesting study on how the resource cost after synthesis is much lower than the analytical LUT cost model we had introduced previously. This opens up more costly design spaces we had not previously explored due to resource constraints. We also delve into how we can design heuristics for LUT Costs and Congestion Estimation and integrate that with the library in the future to discover topologies that push the gap between the analytical LUT estimate and the actual LUT cost further, and lends us with a less congested neural network. While we are unable to propose methods to integrate congestion estimation, we do give insight into how a future heuristic that aids LUT cost reduction could look like.\\
Chapter 6 focuses heavily on implementing LogicNet in a real world scenario. We choose Jet Substructure Classification as our target problem, as they have data-rates in excess of hundreds of terabytes per second and require triggers that have very low target latency and use custom hardware like FPGAs or ASICs. We are able to discover several topologies that give very good performance with a very low analytical LUT cost. Further, we gain insight into the important of SoftMax in the Jet Substructure Classification problem, and do some topology exploration. We also compare the accuracy of models pruned by different methods.\\
Chapter 7 introduces the well known data-set, MNIST. We provide the reader with insights into the behavior of LogicNet on this dataset for both MLPs and Convolutional Neural Networks. We study the interplay of Analytical LUT cost and accuracy, as well as the relation between accuracy and the bit-width. We study heterogeneous architectures with 'skip connections' for both MLPs and Convolution, and find some benefit to it. For MLPs, it is interesting to note that we gain benefits with skip connections without increase in hardware cost. We do not discuss the Place and Route effects of introducing skip connections in this thesis. \\
We believe that there are an array of problems which could find great use for the LogicNet library, and this thesis hopes to be of use when making architectural decisions. We hope to further continue this work and learn more about how to discover sparse topologies, and drive down the resource cost of such mapping of neural networks to an FPGA fabric.




\begin{fullwidth}
\printbibliography{}
\end{fullwidth}

\cleardoublepage{}





\end{document}